\documentclass[11pt]{article}

\usepackage{amsmath}
\usepackage{natbib}
\usepackage{longtable}
\usepackage{graphicx}
\usepackage{booktabs}
\usepackage{tabularx}
\usepackage{ltablex}
\usepackage{authblk}
\usepackage[hidelinks]{hyperref}
\keepXColumns

\title{Evolving Machine Learning in Non-Stationary Environments: A Unified Survey of Drift, Forgetting, and Adaptation}

\author[1]{Ignacio Cabrera Martin}
\author[2]{Subhaditya Mukherjee}
\author[1]{Almas Baimagambetov}
\author[2]{Joaquin Vanschoren}
\author[1]{Nikolaos Polatidis}

\affil[1]{School of Architecture, Technology and Engineering, University of Brighton, United Kingdom}
\affil[2]{Department of Mathematics and Computer Science, Eindhoven University of Technology, Netherlands}

\date{}

\begin{document}

\maketitle
\begin{abstract}
In an era defined by rapid data evolution, traditional Machine Learning (ML) models often struggle to adapt to dynamic environments. Evolving Machine Learning (EML) has emerged as a pivotal paradigm, enabling continuous learning and real-time adaptation to streaming data. While prior surveys have examined individual components of evolving learning — such as drift detection—there remains a lack of a unified analysis of its major challenges.
This survey provides a comprehensive overview of EML, focusing on four core challenges: data drift, concept drift, catastrophic forgetting, and skewed learning. We systematically review over 100 studies, categorizing state-of-the-art methods across supervised, unsupervised, and semi-supervised learning. The survey further explores evaluation metrics, benchmark datasets, and real-world applications, offering a comparative perspective on the effectiveness and limitations of current approaches and proposing a taxonomy to organize them.
In addition, we highlight the growing role of adaptive neural architectures, meta-learning, and ensemble strategies in managing evolving data complexities. By synthesizing insights from recent literature, this work not only maps the current landscape of EML but also identifies key research gaps and emerging opportunities. Our findings aim to guide researchers and practitioners in developing robust, ethical, and scalable EML systems for real-world deployment.
\end{abstract}

\paragraph{Keywords.} Evolving machine learning, Artificial Intelligence, Survey, Concept drift, Data drift, Catastrophic forgetting, Skewed learning

\section{Introduction}

The proliferation of data from IoT, social media, and networks has revolutionized industries like finance, healthcare, and autonomous systems, generating dynamic data streams that challenge traditional ML models. These models, rely on static data and struggle to adapt to evolving distributions, leading to performance degradation. For instance, in fraud detection, evolving tactics cost businesses billions annually due to undetected patterns, while in healthcare, shifting patient data risks misdiagnoses, impacting patient outcomes. Scientifically, non-stationary environments demand models that learn incrementally, addressing data drift, concept drift, and catastrophic forgetting. EML meets these challenges by enabling real-time adaptation, ensuring robustness and scalability in dynamic settings.

Despite the success of modern machine learning systems, all  paradigms are still based on the assumption of a static environment. Models are typically trained once on a fixed dataset obtained from a stationary distribution and then deployed with the expectation that future data will closely resemble the training data. In reality, nearly every real-world application encounters non-stationary conditions. Data distributions shift gradually or abruptly through concept drift and covariate shift, user behavior evolves, new classes appear or vanish, sensors degrade, and even the target task can change over time. The research community has developed numerous specialized approaches to tackle aspects of this challenge, including continual learning, lifelong learning, incremental learning, online learning, domain adaptation, transfer learning under distribution shift, and various drift detection and handling techniques. However, these efforts remain scattered across different sub-fields that employ different terminology, benchmarks, and core assumptions, which prevents the emergence of a unified theoretical foundation and broadly applicable solutions. To bring together these closely related concepts, namely the management of non-stationary data, the necessity of continuous adaptation, the structural and parametric evolution of models, and the long-term goal of autonomous improvement in open-ended environments, we introduce the unifying term EML. EML moves us away from models that just react when things change, and instead builds systems that actively keep improving their knowledge, structure, and way of learning for their entire life. This takes machine learning from "train once and hope it still works later" to truly smart, ever-adapting intelligence.

From a business perspective, EML reduces costs by automating real-time adaptation, e.g., saving millions in fraud detection by catching evolving patterns. Scientifically, EML advances continual learning, addressing non-stationary data challenges. For policymakers, EML aligns with the EU AI Act \citep{ref118}, ensuring ethical deployment through fairness and transparency. Theoretically, EML contributes to lifelong learning frameworks, while managerially, it enables scalable, autonomous systems for dynamic industries.

In contrast, Online Learning has gained traction as a paradigm where models are continuously updated with new data, enabling them to adapt their predictions in real time. Traditional ML typically requires collecting all data and labels before training. Once deployed, their parameters remain fixed unless a new version is trained from scratch—an approach poorly suited to real-world, evolving scenarios. EML extends Online Learning by incorporating methods that specifically address dynamic data distributions, drift, and adaptation challenges, making it increasingly relevant in industry and research.

By incorporating new data incrementally, Online Learning reinforces predictive power and often improves model performance. The exponential rise of the Internet of Things (IoT) has further accelerated the need for EML, as connected devices now generate massive real-time data streams that must be processed with minimal delay. This survey reviews these techniques in detail and analyses their suitability for different categories of tasks based on specific attributes.

Beyond technical advances, the growing role of AI in high-stakes applications such as fraud detection and predictive maintenance has driven policymakers to engage with AI experts worldwide. Their goal is to establish frameworks that ensure AI development aligns with ethical principles. For example, the EU AI Act \citep{ref118}, the UK Parliament’s Research briefing \citep{ref15}, and the Guidelines for Trustworthy AI \citep{ref47} emphasize requirements such as human oversight, robustness, privacy and data governance, transparency, fairness, societal well-being, and accountability. This survey considers these factors while discussing methods and their deployment.

As with any approach, Online Learning and EML face significant challenges, including data and concept drift, catastrophic forgetting, and skewed learning. While prior surveys address individual aspects of evolving learning-such as drift detection or continual learning-there is no unified analysis integrating all four core challenges under a single EML framework \citep{ref71,ref56}, but a holistic analysis of EML’s core challenges, with emphasis on efficiency and ethics, is lacking. This survey aims to stablish a unified taxonomy and analytical framework that connects these challenges to methodological strategies and evaluation criteria, as well as the challenges that remain open. Unlike existing surveys on drift detection, class imbalance, or continual learning, this work provides a holistic taxonomy across these challenges, integrates ethical and regulatory aspects, and synthesizes case studies, making it the first comprehensive overview of EML as an overarching paradigm, and our analysis reveals that adaptive ensembles and meta-learning strategies consistently outperform static architectures under combined data and concept drift, yet lack standardised benchmarks for cross-domain validation.

The main contributions of this survey are as follows:

\begin{enumerate}
\item[-] We derive a taxonomy linking detection, mitigation, and adaptation strategies across data drift, concept drift, catastrophic forgetting, skewed learning, and network adaptation, revealing interdependencies often overlooked in prior literature.

\item[-] We propose a detailed classification of methods based on their suitability for different challenges, and summarize relevant datasets, models, metrics and future research directions.
\end{enumerate}

This paper is organized as follows: Section 2 delivers the survey methodology, Section 3 reviews the background and key characteristics of EML, Section 4 provides the classification of the methods based on the challenge that they mitigate, Section 5 analyses recent methods across key areas, Section 6 outlines future research directions, and Section 7 concludes with a summary of findings.

\section{Survey methodology}

To ensure rigor and transparency, we adopted a structured review protocol using the PRISMA guidelines as follows:

\textbf{Databases searched}: We queried major scientific databases including Scopus, IEEE Xplore, ACM Digital Library, SpringerLink, and Web of Science, as these provide broad coverage of AI and machine learning research.

\textbf{Search strategy}: The search combined keywords such as “evolving machine learning”, “concept drift”, “catastrophic forgetting”, “online learning”, “adaptive neural networks”, and “class imbalance streams”. Boolean operators were used to expand or refine the queries.

\textbf{Time frame}: The review focused on studies published between 2018 and 2024, capturing the recent wave of research in evolving and continual learning. Earlier works were included selectively when they represent seminal contributions (e.g., ADWIN for drift detection, EWC for catastrophic forgetting). To ensure recency, we focus primarily on works published after 2015, and especially 2018–2024 when interest in evolving ML accelerated. Earlier works are cited only when they represent seminal contributions that underpin current research (e.g., ADWIN for drift detection, EWC for catastrophic forgetting, SMOTE for imbalanced learning).

\textbf{Inclusion criteria}: Peer-reviewed journal or conference publications, papers proposing new algorithms, frameworks, or benchmarks for evolving ML and surveys or comparative studies addressing adaptation in non-stationary environments.

\textbf{Exclusion criteria}: Studies without empirical validation, works limited to static ML techniques and non-English or duplicate publications.

\textbf{Screening and selection}: The initial queries returned several hundred papers. After applying the criteria above, we retained a set of the most relevant works that span the five core challenges of evolving ML: data drift, concept drift, catastrophic forgetting, skewed learning, and network adaptation. These form the basis of the taxonomy and synthesis presented in this survey.

\section{Background Of Evolving Machine Learning}

This section provides the background on EML by covering its definition and the key characteristics, its relationship with traditional machine learning, along with its differences, and the unique aspects of evaluation in these kinds of systems.

\subsection{Definition And Characteristic}

EML is a typical example of AI that involves the incremental and continuous learning process, allowing the models to adapt to data changes and shifting environments, where it has been proven that the EML methods and approaches can be used in all three parts of ML, at the preprocessing stage (feature selection and resampling methods), learning stage (parameter setting, membership functions, and neural network topologies adaptations), and during postprocessing (rule optimisation, decision tree/ support vector pruning, and ensemble learning). The core characteristics of EML systems contribute significantly to their effectiveness and suitability for dynamic environments. Thus, this ongoing adaptation allows EML systems to quickly respond to emerging patterns, thereby maintaining performance levels without substantial degradation.

One crucial aspect of EML systems is incremental learning. As new data is introduced, incremental learning enables the models to assimilate new patterns while preserving the valuable insights acquired from previously processed data. This gradual learning process significantly reduces the retraining costs and optimizes computational resources, ensuring efficiency over time. Additionally, autonomy and self-regulation further enhance the utility of EML systems. These models are capable of autonomously detecting shifts and changes within the data environment, negating the need for continuous human oversight. Through internal mechanisms of self-regulation, the models can monitor their performance, swiftly identify deviations or anomalies, and automatically initiate corrective measures. This feature makes them particularly advantageous for deployment in large-scale or distributed settings. Moreover, real-time operation is integral to the design of EML systems. These algorithms are structured to manage continuous streams of incoming data effectively. Their ability to immediately respond to critical changes ensures sustained operational performance and reliability, making them ideal for scenarios where timely reactions are essential.

Finally, scalability remains a vital characteristic of EML systems. To maintain optimal effectiveness, these models dynamically alter their structure, scaling efficiently in response to evolving data demands. This inherent scalability ensures that EML systems remain robust and effective, even as data volumes and complexities grow.

\subsection{Relationship With Traditional Machine Learning}

ML models aim to provide the predictive capabilities and automatization of repetitive tasks that involve data analysis and decision-making, it is relatively normal to say that traditional ML and EML share the foundational principles and methodologies. In some cases, EML can be viewed as an extension and enhancement of traditional ML, where they are often built upon foundational algorithms such as Decision Trees, Support Vector Machine (SVM) or neural networks, and simply address the limitations that traditional ML poses with dynamic and non-stationary data streams.  Both methodologies (ML and EML) share significant similarities, primarily as they are developed for the discovery of meaningful patterns within data sets. They strive for robustness by optimizing predictive accuracy and simultaneously minimizing prediction errors, thereby ensuring reliable outcomes. Furthermore, these approaches significantly contribute to enhancing decision-making capabilities by effectively integrating and analysing the available data. Central to both methodologies are tasks involving classification and regression, highlighting their versatility and broad applicability across various analytical scenarios.

In both methodologies, supervised and unsupervised, learning techniques can be developed and used, reflecting their comprehensive nature and flexibility in addressing diverse analytical problems. Additionally, they utilize sophisticated feature engineering techniques such as normalization, encoding of categorical features, and dimensionality reduction, all essential for improving the overall efficiency and accuracy of data models. Reliance on optimization methods such as gradient descent or Adam, alongside regularisation techniques like dropout or batch normalization, underscores their commitment to model stability and accuracy. Lastly, both methodologies employ crucial evaluation metrics, including Accuracy, Precision, Recall, F1-score, Mean Absolute Error (MAE), Mean Squared Error (MSE), and ROC curves, to rigorously assess and ensure the effectiveness of the developed models.

On the other hand, the primary distinction between evolving machine learning (EML) and traditional machine learning (ML) lies in adaptability. EML continuously adjusts as data evolves, dynamically responding to changes. In contrast, traditional ML systems exhibit limited adaptability, typically requiring periodic updates to remain effective, while feature extraction and selection methods remain critical within the EML context. However, these methods have evolved significantly to operate dynamically, continuously adapting to incremental data streams. This ongoing adjustment capability ensures that EML systems remain relevant and effective even as data characteristics shift.

Regarding evaluation metrics, EML introduces specialized measures tailored to the unique challenges of dynamic data streams. These new metrics specifically quantify aspects such as the degree of data change, including indicators like forgetting rate and adaptability scores. These metrics provide essential insights into how effectively an EML system adapts over time.

Another advantage of EML systems is their reduced maintenance requirement. Due to their inherent adaptive nature, these systems can autonomously adjust, minimizing the need for manual interventions and updates. Consequently, EML systems offer efficiency benefits through lower maintenance efforts compared to traditional ML.

Moreover, robustness is another notable strength of EML systems, as they are explicitly designed to detect and adapt to data drifts effectively. This feature significantly enhances their reliability in environments characterized by frequent or unpredictable data changes.

Finally, catastrophic forgetting, a phenomenon where previously learned information is lost when new data is incorporated, is generally not a concern in traditional ML due to its static data approach. However, managing catastrophic forgetting is a critical aspect of EML, as systems must retain previously acquired knowledge while adapting to new information streams.

\subsection{Evaluation In Evolving Machine Learning}

Since evolving ML deals with changing environments, we need dynamic metrics that assess performance over time. In this kind of scenario, traditional static metrics are not as effective as they do not consider the different challenges associated with continuous learning, and it is required to use more specialised evaluation metrics that capture the model’s accuracy and adaptability over time, and this is the case of dynamic techniques and metrics.

\subsubsection{Meta-Learning and Transfer-Based Approaches}
When referring to techniques, the Prequential Evaluation, specifically the Interleaved Test-Then-Train approach, is a sequential evaluation technique that involves alternating between testing and training phases for each data instance. This method assumes that every individual instance initially serves to evaluate the model's accuracy and subsequently updates the model through training, aiming to progressively enhance model accuracy and efficiently utilize available data. The prequential error, a crucial component of this evaluation, is computed cumulatively through a loss function that compares predicted values to actual observed outcomes, formally represented in Equation \ref{Equation 1}, where it takes a sequence of  observations and  represents the loss function used for the type of problem, taking  as the prediction made by the model before seeing the true label :

\begin{equation}
\label{Equation 1}
\textit{Prequential Error} = \sum_{t=1}^{n} f(\hat{y}_t, y_t)
\end{equation}

There are three primary prequential evaluation techniques commonly implemented across various approaches, each addressing specific scenarios and data characteristics: Landmark Window (Interleaved Test-then-Train), Sliding Windowing, and Forgetting Mechanisms. Each of these methods offers a unique strategy for managing data streams and dealing with evolving data distributions, reflecting different perspectives on how past data relevance should diminish over time. Additionally, the Controlled Permutations evaluation approach addresses the challenge of accurately assessing adaptive learning algorithms. This method reduces the masking of adaptive properties—common when averaging performances across various algorithms—by running multiple tests on randomly permuted versions of data streams. By preserving different distributions within these permutations, this technique effectively measures model robustness and adaptability, particularly beneficial in scenarios involving sudden concept drifts. Controlled permutations create diverse test sets, enabling the evaluation of model volatility, robustness, and parameter optimization, thus minimizing overfitting risks associated with sequential data ordering.

Lastly, the concept of the 'illusion of progress' highlights a prevalent issue in the field: the tendency to demonstrate methodological advancements primarily through the introduction of new classifiers and reliance on accuracy metrics. Some critiques this practice, suggesting a broader evaluation methodology that encompasses more than just classifier performance, where they emphasize the necessity of expanding evaluation criteria beyond accuracy alone to gain a more comprehensive understanding of model performance and methodological advancements.

In addition to that, several critical metrics have been identified to assess model performance effectively before and after drift events. These include Drift Detection Delay, or Latency, which measures the elapsed time between the occurrence of a drift and its detection by the monitoring system. Minimizing this delay is essential to enable rapid response to shift in data distribution, with the calculation explicitly defined as the difference between detection and drift occurrence times as shown in Equation \ref{Equation 2}, where  is the time when the drift detection method signals the drift, and  is the time when the drift actually occurs.

\begin{equation}
\label{Equation 2}
D_{delay} = t_{detection} - t_{drift}
\end{equation}

The Magnitude of the drift aims to calculate the distance between two concepts and quantify the severity of it, like Latency, but in this case, they do not specify the measure of distance to be used, identified as $D$, where $t$ and $u$ identify the concepts in different times. The formula can be seen in Equation \ref{Equation 3}.

\begin{equation}
\label{Equation 3}
Magnitude_{t,u} = D(t, u)
\end{equation}

Another metric is the Duration of the drift measuring the elapsed time over which a period of drifts occurs. The formula shown in Equation \ref{Equation 4} aims to calculate this duration based on the starting time $t$, and the ending time $u$.

\begin{equation}
\label{Equation 4}
Duration_{t,u} = u - t
\end{equation}

A statistical measure called the Hellinger Distance, is widely used to quantify the similarity or difference between two probability distributions, in a range between 0 and 1, where 0 indicates perfect similarity and 1 maximum dissimilarity. The formula is presented in Equation \ref{Equation 5}, where the $P$ and $Q$ values represent the probability distributions to measure, then  and  are the probabilities of the $i^{th}$ event in the P and Q distributions.

\begin{equation}
\label{Equation 5}
H(P, Q) = \frac{1}{\sqrt{2}} \sqrt{ \sum_{i=1}^{n} \left( \sqrt{p_i} - \sqrt{q_i} \right) }
\end{equation}

The Path Length aims to calculate the path that the drift traverses during a period of drift. It is defined as the cumulative deviation observed during that period. The formula is bounded with the Drift Magnitude as can be seen in Equation \ref{Equation 6}, where $D$ represents the distance between data distributions at two consecutive time steps,  and , and $t$ represents the time interval of these distributions.

\begin{equation}
\label{Equation 6}
PathLen = \sum_{i=1}^{n} D(P_{t_i}, P_{t_{i-1}})
\end{equation}

A critical measure that helps to quantify the velocity that the distribution is changing at a given time $t$ is the Drift Rate, where the base formula can be seen in Equation \ref{Equation 7}, where given two distributions,  and , at a given time $t$, the distance metric is defined as $D$ and the interval between the two points is provided as .

\begin{equation}
\label{Equation 7}
Drift\ Rate = \frac{D(P_{t+\Delta t}, P_t)}{\Delta t}
\end{equation}

This is further simplified into an Average Drift Rate by integrating Path Length over the drift duration, where  and  refer to the start and end times, providing a comprehensive view of drift dynamics as can be seen in Equation \ref{Equation 8}.

\begin{equation}
\label{Equation 8}
Avg\ Drift\ Rate = \frac{PathLen}{t_2 - t_1}
\end{equation}

On the other hand, the Local Drift Degree (LDD) helps to quantify regional density discrepancies between two different sample sets, thereby, identifying density increased, decreased and stable regions. Unlike global drift measures, which assess overall distribution changes over larger periods, LDD specifically evaluates short-term, localized shifts, making it suitable for precise and timely drift detection. The formula commonly used can be defined as the distance between two probabilities, and often it is expressed using measures such as Kullback-Leibler divergence, Jensen-Shannon divergence, or statistical Kolmogorov-Smirnov (KS). It is globally expressed as seen in Equation \ref{Equation 9}.

\begin{equation}
\label{Equation 9}
\delta_W = \frac{\dfrac{|B_W|}{n_b}}{\dfrac{|A_W|}{n_a}} - 1
\end{equation}

Denote the feature space as $W$, where in an ideal situation $|A_W|$ and $|B_W|$ represent the number of data instances in $W$ that belong to $A, B$, and ,  represent the total number of instances in $A$ and $B$ separately.

Similarly, the Margin Density (MD) assesses the density of data points near the decision boundary, also known as the margin, where the classifier is uncertain and tries to guess based on its learned information. The higher the density is presented indicates that the model is struggling to correctly classify incoming data, while the lower the density is, a clearer separation between classes is presented, resulting in the model correctly classifying the new incoming data. The formula for a given margin  can be seen in Equation \ref{Equation 10}.

\begin{equation}
\label{Equation 10}
MD(\tau) = \frac{\text{Number of data points within margin } \tau}{\text{Total number of points}}
\end{equation}

The Conditioned Marginal Covariate Drift is the weighted sum of the distances between each of the conditional probability distributions, given as $P$, over the possible values of the covariate attributes for each specific class determined by $y$, from a period from $t$ to $u$, where the weights are the average probability of the class over the two time periods. The formula can be seen in Equation \ref{Equation 11}.

\begin{equation}
\label{Equation 11}
\sigma_{t,u}^{X|Y} = 
\sum_{y \in Y} 
\left[
\frac{P_t(y) + P_u(y)}{2} 
\cdot 
\frac{1}{2}
\sum_{x \in X} 
\left| P_t(\tilde{x} \mid y) - P_u(\tilde{x} \mid y) \right|
\right]
\end{equation}

In the same study, the Posterior Drift is identified where for each subset of the covariate attributes there will be a probability distribution over the class labels for each combination of covariate values,  at each period. This can be calculated as the weighted sum of the distances between these probability distributions, where the weights are the average probability over the two periods of the specific value for the covariate attribute subset as seen in Equation \ref{Equation 12}.

\begin{equation}
\label{Equation 12}
\sigma_{t,u}^{Y|X} = 
\sum_{y \in X} 
\left[
\frac{P_t(\tilde{x}) + P_u(\tilde{x})}{2} 
\cdot 
\frac{1}{2}
\sum_{y \in Y} 
\left| P_t(y \mid \tilde{x}) - P_u(y \mid \tilde{x}) \right|
\right]
\end{equation}

The study from Anderson et al. \citep{ref5} introduces the Lift-per-Drift (LPD) aiming to measure drift detection performance through its impact on classification accuracy and penalized by drifts detected on a dataset, more specifically, it measures the improvement in predictive performance, considered “lift”, that is achieved by a model after detecting and adapting to a concept drift event. It is based on a performance metric, such as accuracy or error rate. The formula can be seen in Equation \ref{Equation 13}, where $n$ indicates the number of detected drift events, and both performances just after and before adaptation following the $i^{th}$ drift detection event.

\begin{equation}
\label{Equation 13}
LPD = \frac{\sum_{i=1}^{n} \left( Performance_{post\text{-}drift_i} - Performance_{pre\text{-}drift_i} \right)}{n}
\end{equation}

Finally, the Novel Precision Rate $N_{prc}$ measures the proportion of correctly identified novel category samples among all samples classified as novel by the model. Equation \ref{Equation 14} presents the corresponding formula, where $T_{nov}$ denotes the number of samples correctly detected as novel, and $T_{nov} + F_{exostnov}$ represents the total number of samples predicted as novel, including both correct and incorrect detections.

\begin{equation}
\label{Equation 14}
N_{Pre} = \frac{T_{nov}}{T_{nov} + F_{existnov}}
\end{equation}

When evaluating algorithmic efficiency, several mathematical benchmark functions are frequently employed. The Rastrigin Function is notable for its numerous local minima, making it ideal for assessing global search capabilities. The Ackley Function, characterized by its flat regions and central hole, poses significant convergence challenges. Additionally, the Rosenbrock Function is particularly valuable for testing optimization on a non-convex, curved surface, whereas the Schwefel Function, with its multiple local minima, provides a rigorous evaluation framework for global optimization algorithms.

Synthetic data streams have also been developed specifically to test algorithm adaptability and performance. The SEA Generator produces concepts featuring incremental drifts, ideal for classification algorithm assessment. Meanwhile, the Hyperplane Generator simulates evolving hyperplane-based concepts, which are essential for adaptive learning evaluation. Other generators include the Circles Generator, designed to create circular class boundaries useful in classification boundary detection, and the SINE Generator, which produces sinusoidal decision boundaries to rigorously test classifier adaptability. The MIXED Generator further enhances testing capabilities by combining multiple concept drifts within data streams.

Benchmarking using real-world datasets remains crucial for realistic and comprehensive algorithm evaluation. The UCI Machine Learning Repository \citep{ref74} provides widely used datasets such as the Wine Quality Dataset \citep{ref84} for regression and classification tasks, the Wisconsin Breast Cancer Dataset \citep{ref109} for binary cancer diagnosis classification, the E-coli Dataset \citep{ref78} for multiclass biological classification tasks, and the Yeast Dataset \citep{ref77}, which challenges classifiers with multiple protein localization classes. Similarly, the KEEL Repository \citep{ref2} offers specialized datasets beneficial for evaluating machine learning algorithms, particularly in evolving feature selection and algorithm benchmarking.

The WEKA Datasets repository is widely integrated into the WEKA \citep{ref51} platform and facilitates convenient benchmarking across various domains, including classification, regression, and clustering tasks. In addition, large-scale and specialized datasets such as the Forest Cover Dataset \citep{ref16}, used for predictive analytics in forest classification, and the Electricity Market Dataset, essential for real-time predictive modelling of electricity price movements, enhance real-world applicability. Other significant datasets include the Human Activity Recognition (HAR) Dataset \citep{ref70} for sensor-based activity classification, the Poker Hand Dataset \citep{ref87} for evaluating algorithms via poker hand recognition, the Airlines Dataset for airline delay predictions, the Air Quality Dataset \citep{ref104} for environmental forecasting, the Rialto Bridge Dataset for anomaly detection in structural health monitoring, and the cybersecurity-focused KDDCup99 Dataset \citep{ref90}.

OpenML \citep{ref103} is an open-source machine learning platform that provides almost 6.4k+ datasets (most of the datasets mentioned in this paper can be easily loaded using the OpenML Python API) in uniform standards along with standardized tasks and workflows to enable reproducibility and transparent research. It tries to future proof the availability of datasets by allowing versioning and allowing users to link their experiments to these datasets through easily reproducible pipelines. AMLB (AutoML Benchmark) \citep{ref39} is a tool that builds on OpenML to facilitate easier benchmarking of models on tabular data. A special tag that links the two are benchmark datasets, a way of stating that a certain dataset is important for a sub class of tasks.

In conclusion, we should mention that in the domain of computer vision, datasets such as MNIST \citep{ref31} and CIFAR \citep{ref63} are extensively used. MNIST serves as the standard benchmark for handwritten digit classification, while CIFAR presents diverse natural image data, challenging object recognition and classification algorithms across multiple categories, making it a common benchmark dataset for any kind of image classification model.

\subsubsection{Limitations of Current Evaluation Metrics}

The observations presented here concern the methodological constraints associated with several evaluation metrics commonly used in EML. They are included at the end of this section because these considerations refine the interpretation of the metrics introduced above, rather than reflecting the intrinsic learning challenges examined in later sections. Their placement here emphasises that the limitations pertain to the design and validity of empirical evaluation protocols rather than to the conceptual taxonomy of EML itself.

Distance based measures such as the Hellinger distance in Eq.~5 rely on discretised approximations of continuous probability distributions. When applied to high dimensional data, histogram based estimates become sparse and unstable, which substantially reduces the reliability of the resulting distance computation. As a consequence, the Hellinger distance is most suitable in low dimensional feature spaces or when used in conjunction with dimensionality reduction techniques that yield compact representations of the data.

Metrics that quantify post drift recovery, such as the Lift per Drift (LPD) measure in Eq.~13, offer a distinct but equally important perspective. Because drift typically induces an immediate decline in model performance, the LPD value reflects the extent to which the model is able to recover through adaptation. It therefore serves as an indicator of the resilience and adaptability of the learning system rather than a direct measure of absolute predictive quality. This distinction is especially relevant in evolving environments where the difficulty or structure of the target concept may change over time, making raw accuracy an incomplete indicator of model competence.

These considerations underscore the importance of selecting evaluation metrics that are well aligned with the statistical and temporal properties of evolving data. Careful interpretation of these measures is essential for drawing valid conclusions about the behaviour of adaptive models in non stationary environments.

\section{EML Challenges}

\begin{figure}[h!]
    \centering
    \includegraphics[width=0.9\textwidth]{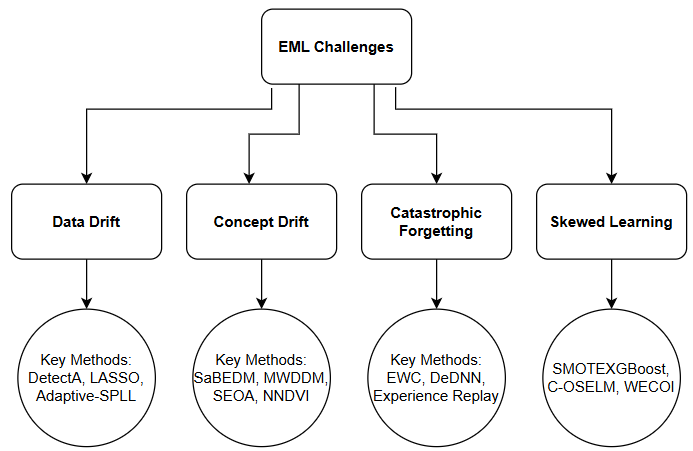}
    \caption{Taxonomy}
    \label{fig:eml_challenges}
\end{figure}

This section provides a deeper understanding of the different challenges involved in EML, that traditional static machine learning systems have not encountered. They can be divided into four main categories based on the extent of the challenge, where we can find Data Drift, Concept Drift, Catastrophic Forgetting, and Skewed Learning. From the extensive literature reviewed for this paper, it has been demonstrated that most current surveys focused on individual challenges, whereas only \citep{ref98} provide a study on all the challenges associated with online learning with full feedback. Figure \ref{fig:eml_challenges} below provides a taxonomy of EML challenges.

\subsection{Data Drift}

This challenge is presented when the statical properties of the input features change over time, as the data that it encounters has changed from the initially used during the training, impacting directly the long-term general performance of the models.

Between the surveys covered for this topic, Palli et al. \citep{ref81} investigated the performance of existing drift detection methods on multi-class imbalanced data streams with different drift types and proposed DDM as the best method based on the average F1 score. Ashok et al. \citep{ref8} compared and evaluated the main approaches in time-series models, providing a taxonomy and in-detail evaluation metrics, and Gemaque et al. \citep{ref38} proposed an overview of unsupervised methods.

\subsubsection{Unsupervised Detection Methods}
Table 1 provides a summary of the methods examined for handling data drifts, where an study proposed the Adaptive-SPLL \citep{ref59} following the suggestion of using K-means clustering for estimating the density of the Gaussian Mixture model and also introduced adaptive window sizes, where the size of the windows grows until a drift is detected, or max size is reached, then it will split continuously into smaller windows to detect the actual starting time of the current drift, reusing the K-means centroids from the last window. This solution has been evaluated using an industrial dataset obtained from the NASA, containing C-MAPSS for degradation, showing an increase in accuracy and reduced computation time, but similar detection delay than non-adaptive solutions.

The DetectA \citep{ref34} proposed is a proactive drift detection approach that identifies abrupt drifts in the data stream before they occur, by using a Multivariate Hypothesis test to compare the statistics that could indicate the presence of drift. Two different experiments were performed, where the first aimed to identify the sensitivity of the approach by computing its false positives and false negatives, where their results suggest that this detector is more efficient for high dimensional datasets with intermediate size blocks, using datasets with any proportion of drifts and imbalanced binary classes. The second one aimed to measure the performance of the detector using a multi-layer perceptron (MLP) as the baseline model along with different datasets, such as SEA concepts, Nebraska, Electricity, Cover Type, and Poker Hand. Different versions of the method are compared, such as no detection, reactive detection, and proactive detection, where the second ones seem to perform better and help to improve the general performance of the model, the no detection versions were faster in computational time as expected.

Other approaches take an adaptive data augmentation approach, such as the LITMUS-adaptive \citep{ref99} that extends the LITMUS-static by applying a detector monitoring when a signal exceeds a given threshold, applied for physical event detection using data for social networks, then a new classifier is trained using the previous data window available, and previous models get their parameters updated. In this method, all the previously created models are stored in a database as key-values of the new and previous model, along with the centroids as the key for a classifier. The prediction is then obtained by producing voting between the ensemble of models. The experimental results showed that the adaptive version outperforms the static one with almost 350\% more landslide events identified, where the f-score reached a 98\% against the 76.2 of the static.

\subsubsection{Supervised Detection Methods}
The LASSO \citep{ref85} proposed a feature rank drift detection procedure based on analysing changes in feature ranks across adjacent chunks of data, and monitoring its fluctuations by comparing them to the reference feature ranks. When a change is identified, an alarm gets triggered and the classifier is retrained on the new data chunk, updating the baseline ranks to reflect the new distribution. Different tests were conducted, the ones using Synthetic showed that the proposed solution outperformed DDM, EDDM, ADWIN, and PCA-FDD for the Abrupt and Recurring data drift types, while the result on the real-world datasets seems to deliver almost as the other compared methods, but lowering considerably the computational time in almost every scenario.

A method that measures the magnitude of image classification tasks on noisy weather effects tasks \citep{ref91} that uses MNIST and CIFAR10 images, and stores the prediction probabilities of each class in the network, helping to identify all kinds of drifts and estimate the magnitude with high accuracy, but has not being compared with other methods. This approach has been extended by Ye et al. \citep{ref112} using cumulative distribution functions (CDFs) of output probabilities instead of base prediction probabilities. In the experiments, the proposed method used CIFAR10 along with a RandomRain function to demonstrate that using the two drift magnitudes as a combined pair reduces the overestimation of drift levels, helping to increase the accuracy.

\subsubsection{Semi-Supervised Detection Methods}
Similarly, Krawczyk et al. \citep{ref62} introduced strategies that use a drift detector module, such as ADWIN2, that uses the labelled samples as the active learning, and when the accuracy of the classifier starts decreasing, another classifier is trained on the background using arriving objects, so in case of the change is finally detected, the classifier is replaced. The Random strategy++ randomly draws instance labels with probability equal to the assumed budget, this is done by improving the label query if a change is detected, resulting in increasing the labelling probability according to the output of the drift detector. The Variable uncertainty strategy++ is based on monitoring the certainty of classifiers, aiming to label the least certain instances within a time interval, and resulting in a balanced budget over the time. In this strategy, the threshold is rapidly reduced once the alarm or the drift is detected to allow the gathering of the highest number of labelled objects to quickly adapt a new model. Finally, the Randomised version of the previous strategy, where the threshold is simply randomly selected. Experiments between the different versions were conducted and compared with a fully supervised classifier using different benchmark datasets, such as Airlines, Electricity, Forest Cover, RBF, and Hyperplanes, and evaluated using prequential accuracy to determine if these strategies are closer in accuracy to the fully labelled classifier, where the results showed that the only in the electricity datasets the proposed active learning strategies were similar, even with reduced budget, while the remaining datasets showed a significant increase in accuracy when the feedback received from the drift detector is utilised by the label query.

\small
\begin{longtable}{@{}>{\raggedright\arraybackslash}p{1.5cm}p{2cm}p{3.2cm}p{1.6cm}p{3cm}@{}}
\caption{Drift detection methods\label{tab:drift_detection}}\\
\toprule
\textbf{Ref} & \textbf{Category} & \textbf{Method} & \textbf{Drift} & \textbf{Datasets} \\
\midrule
\endfirsthead

\caption*{\textit{Table \thetable{} (continued)}}\\
\toprule
\textbf{Ref} & \textbf{Category} & \textbf{Method} & \textbf{Drift} & \textbf{Datasets} \\
\midrule
\endhead

\bottomrule
\endfoot

\citep{ref59}  & Unsupervised & Adaptive SPLL & N/A & Synthetic in-house dataset for lab testing and NASA data with C-MAPSS for degradation \\
{\citep{ref34}}  & Unsupervised & DetectA & Abrupt & SEA Concepts, Nebraska, Electricity, Cover Type and Poker hand \\
{\citep{ref99}}  & Unsupervised & LITMUS-adaptive & N/A & Not specified \\
{\citep{ref85}}  & Supervised & Least Absolute Shrinkage and Selection Operator (LASSO) & All & Synthetic (MOA generated), Phishing, Electricity, Ozone, Arrhythmia, Poker hand, Rialto Bridge timelapse, Outdoor \\
{\citep{ref91}}  & Supervised & N/A & Simultaneous & MNIST, CIFAR10 \\
{\citep{ref112}} & Supervised & N/A & Simultaneous & Social-sensor data (2014--2018; Facebook, Twitter), Rainfall from NOAA, Earthquake data from USGS, Landslide predictions from NOAA and USGS \\
{\citep{ref62}}  & Semi-supervised & RAND++, VAR-UN++, R-VAR-UN++ & All & Airlines, Electricity, Forest Cover, RBF, Hyperplanes, Tree \\

\end{longtable}

\subsubsection{Key Insights}

The methods examined in this section highlight a shift from purely reactive drift detection toward more anticipatory and robust strategies. Approaches such as \textbf{DetectA} exemplify proactive monitoring by examining changes in the input distribution, while \textbf{DDM} remains representative of traditional error driven detection. Techniques based on feature importance, such as \textbf{LASSO}, provide a complementary perspective by identifying structural changes in $P(Y\mid X)$ that are less sensitive to noise. Methods including \textbf{Adaptive SPLL} further illustrate the value of flexible windowing in industrial contexts in which drift often unfolds gradually. Together, these methods suggest that effective data drift detection requires coordinated sensitivity to both input variation and concept level change, supported by mechanisms that remain stable under noise and gradual evolution.

\subsection{Concept Drift}

This type of challenge represents the changes in the underlying relationship between input and target variables, changing the conditional distribution of target variables given the input data. 

\subsubsection{Types of Drift}
\begin{figure}[h!]
    \centering
    \includegraphics[width=0.8\textwidth]{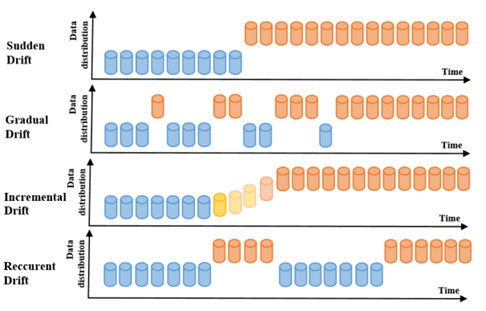}
    \caption{Types of concept drift}
    \label{fig:drift_types}
\end{figure}

Before exploring drift in detail, the types of drift are outlined below. Figure \ref{fig:drift_types} shows the different types of concept drift.
Sudden Drift: It is the shift or change in the data distribution that has occurred abruptly and unexpectedly. In general, it can be disruptive and challenging to detect. Eg: A fraud detection model fails after an unexpected policy change causes a spike in new fraud tactics overnight.

\textbf{Gradual Drift}: This kind of shift or change happens over an extended period, and its effects cannot be immediately noticeable. Eg: Fraudsters slowly adapt their techniques over months, subtly reducing the model’s effectiveness without immediate signs.

\textbf{Incremental Drift}: It is the change in the relationship between the input and the output that happens slowly and gradually. Eg: The types of fraudulent behaviour slowly evolve with minor adjustments, gradually changing the relationship between transaction features and fraud labels.

\textbf{Recurring Drift}: This type of change tends to appear and disappear cyclically, in a patterned way, resulting in periodical changes that are not constant. Eg: Fraud patterns spike during major shopping seasons like Black Friday each year, following a predictable cycle.

\subsubsection{Comprehensive Survey Reviews and Taxonomies}
A wide range of survey reviews have been covered during the production of this study, where it is necessary to mention the work effectuated by Barros and Santos \citep{ref12,ref13} with more than 50 method covered and produce a large-scale comparison, Iwashita and Papa \citep{ref54} with 71 reviews covering methods and public available datasets, Lu et al. \citep{ref71} introduced an extensive survey review with over 130 methods covering all areas of research, 10 synthetic datasets, and 14 public available dataset, Elkhawaga et al. \citep{ref33} also produced an extensive literature review in concept drift analysis and proposed the framework called Concept Drift Analysis in Process Mining (CONDA-PM) that aims to guide users in evaluating the maturity of an approach. Moleda et al. \citep{ref76} analysed methods, techniques, and approaches applied in coal-fired power plants for fault prediction.

Bayram et al. \citep{ref14} proposed a taxonomy classification for all drift detectors that is quite complete and well-presented. Hinder et al. \citep{ref48} presented a comprehensive review of metric choices for concept drift detection, comparing the existing ones and proposing new choices demonstrating their suitability. Hinder et al. \citep{ref49} provide a detailed survey of unsupervised methods, providing a taxonomy of approaches and detailed mathematical definitions. Sakurai et al. \citep{ref89} survey is oriented to guide in the selection of the best concept drift detector by testing and benchmarking them in their work. Suárez-Cetrulo et al. \citep{ref97} provides an extensive survey of supervised and unsupervised detectors that covers passive and active detectors, but also adapters. Xiang et al. \citep{ref110} summarized concept drift adaptation methods for deep learning frameworks. Hovakimyan and Bravo \citep{ref50} proposed a literature review that adheres to PRISMA guidelines aiming to provide a valuable resource for future research.

Werner et al. \citep{ref108} focus on the performance of unsupervised concept drift detectors, their computational complexity, and strategies for benchmarking them. Costa et al. \citep{ref28} focused on the impact of four descriptors (severity, recurrence, frequency, and speed) and conducted experiments on five benchmark datasets with 32 descriptor variations, resulting in a total of 1440 combinations. Lukats et al. \citep{ref72} explore the availability of fully unsupervised concept drift detectors by analysing the performance of ten of them in real-world data streams.

\subsubsection{Sliding Window and Statistical Comparison Methods}
Table 2 summarizes the widely used type of methods to mitigate this issue, where the commonly approach is single or multiple sliding windows are applied to identify subsets of data from continuous data streams, moving forward as the new data arrives, helping the model to compare the statistics received in two or more different data moments and adapt quickly to new conditions. Between these sliding window methodologies, Monte Carlo Dropout in \citep{ref10} under the Uncertainty Drift Detector (UDD) method without relying on true labels, has showed that this application of ADWIN can increase considerably the accuracy, in comparison with KSADWIN and its base version, when evaluated using a multiple benchmark datasets, and metrics such as Mean time Detection (MTD), missed detection count (MDC), Root Mean Squared Error (RMSE), and Matthews Correlation Coefficient (MCC).

The double-window mechanism to handle concept drifting data streams  that calculates the standard deviation and the error rate on the current and previous windows, where other calculations have been applied between these windows, such as the neural network combined approach \citep{ref93} capable of distinguishing between known labelled concept drift and novelty, employing PCA and radial distance, has been evaluated using synthetic and CIFAR10 datasets, compared with CM, MMD, KDQ and LDD by applying metrics such as number of drifts and novelty detected, and Maximum Mean Discrepancy (MMD), Local drift degree (LDD), Accuracy, running time.

The least-squares density-difference estimation approach with its two extensions for sliding window and ensemble approach types \citep{ref17} that is compared with KNN, H-ICI CDT, and CPM tests in unspecified datasets by applying False Positive rate, False negative rate, Delay, and CT execution time, to determine that this solution works in terms of promptness and accuracy.

The Fisher’s Exact test approach \citep{ref19}, based on STEPD, introduces three different methods referred as FPDD, FSDD, FTDD, that were compared using 24 synthetic dataset with DDM, ECDD, SEED, FHDDM, and STEPD using accuracy, precision, Recall, and F1-score, to determine that these methods deliver higher predictive accuracy in most datasets, or in the top 3 with other competitors.

\subsubsection{Distance and Divergence-Based Approaches}
The plover algorithm utilizing Statistical Moments and the Power Spectrum to compare the difference between the data windows \citep{ref75}, that has been evaluated on different synthetic, S\&P500, Electricity, and Airlines datasets by using Standard deviation and Power Spectrum as measures, and compared with CUSUM, PHT, EWMA, SONDE, GWR, and ADWIN, showing that this method can be compared in performance with the ADWIN method.

The D3 method employing the KL divergence to measure the change between two windows \citep{ref40} has been evaluated on Electricity, Covertype, Poker hand, Rialto, Rotating Hyperplane, Moving RBF, and Interchanging RBF by using accuracy as the chosen metric and comparing the results with ADWIN, DDM, and EDDM,  determining that it outperforms the baselines, yielding models with higher  performances on both real-world and synthetic datasets.

The label-less drift detection (L-CODE) \citep{ref116} analysing changes in the space of Shapley values and the feature distribution, has been evaluated on SEA, Electricity, Phishing, Adult, and Rabobank datasets using Accuracy, Feature distribution, and Shapley distribution, against HDDDM, MD3, ADWIN, DDM, and PH, along with NB and HT as base classifiers, proving that the method works but does not outperforms other methods that require labelling.

While another method applying the Bhattacharyya distance metric (BDDM) \citep{ref9} measuring the similarities between probability distributions over time, makes use of Agrawal, LED, RBF, Electricity, Shuttle, Connect-4, Diabetes, Airlines, Poker hand, Covertype, Pen digits, Gesture and Codrna, and the metrics Bhattacharyya distance, Accuracy, precision, recall, FN, FP during evaluation, where the results obtained has been compared to DDM, ECDD, SEED, STEPD, FHDDM, WSTD, applying the Hoeffding Tree or NB as base classifiers, proving that accuracies were comparatively better than those of the other method, with slightly better results in abrupt scenarios than the gradual ones.

The combination of applying Exponentially Weighted Moving Average (EWMA) and the Jensen-Shannon Divergence \citep{ref35}, referred as CDDDE, proposes the calculation of drifts dynamically without the use of labels. This approach has been evaluated on multitude of synthetic and real-world datasets using Accuracy as their sole metric, determining that this approach efficiently detect the concept drift, and the retrained classifier effectively improves the classification accuracy.

\subsubsection{Adaptive Windowing and Weighted Models}
The Sliding Adaptive Beta Distribution Model (SABeDM) \citep{ref6} combines sliding windows of different sizes with the beta distribution technique to track changes in underlying distributions of the data. It can be applied for online learning, data stream mining, and real-time monitoring systems, where it has been compared to SRP, ADWIN, DDM, and EDDM using metrics such as Accuracy, precision, recall, and F1-score, and determined that it offers an effective and fast way to identify concept drifts.

The Multi-level Weighted Drift Detection Method (MWDDM) \citep{ref26} introduced three levels of detection based on the predictions received, the “stable level” occurs when no drift is presented and uses a linear weighting method that aims to assign more weight to new classes and balancing them with the previous ones while updating the model. The “warning level” seeks to increase the difference in weight values between the instances in the long and short windows, and update them, to reduce the drift to stop it happening. Then the “drift level” is determined by the implementation of Hoeffding inequality or McDiarmid inequality tests, depending on the variant of the method, that is aimed to calculate the difference between both windows’ average of correct predictions, highlighting the drift and resetting the classifier for the new data distribution. This approach has been evaluated on SINE, MIXED, CIRCLES, LED, Elec2, Forest Covertype, and Poker hand using Accuracy, Detection delay, TP ratio, FP ratio, and FN ratio, where the results were compared with DDM,  EDDM,  RDDM,  FHDDM,  FHDDMS,  MDDM,  and  HDDM, determining that this method detects drift faster, maintaining a low FP and FN ratio, keeping a high accuracy with lower computational cost.

In addition, multiple small-scale windows incorporating fast variance estimation strategies and variance-based dynamic weighting \citep{ref46} to adjust the weighting of the window data instances, amplifying the window mean and enhancing adaptability, in combination with the variance feedback strategy, which allows providing different mean calculation and statistical testing methods based on different variances, helping to provide a tailored to the data solution. This method has been evaluated on Sine, Mixed, Circles, Led, Elec2, Forest Covertype, Poker hand, and spam datasets using metrics such as TP, FP, FN, detection delay (DD), and average accuracy, comparing the results with FHDDM, MWDDM\_H, MDDM, WMDDM, BDDM, RDDM, DDM, HDDM\_A, and HDDM\_W, proving that it can effectively contend with different types of concept drift and has good robustness and generalization.

\subsubsection{Label-Free and Unsupervised Drift Detection Techniques}
Other methods demonstrated the capability of current techniques for the detection of the changes, such is the case for, the UCDD method \citep{ref92} implying a clustering technique to label the data, and identify points located on the boundaries of the labelled data. The evaluation has been performed on synthetic datasets using drift magnitude as the sole metric, where the results are compared with CM and KDQ, determining that it outperforms the other compared methods.

The model-independent method for image classification \citep{ref96} that converts the data contained on the windows into a 2-D representation and compares the difference between the pixel intensity of them, it has been evaluated on Heartbeats, Flying insects, Sensor Posture, StarLightCurves, UWaveGestureLibrary, and Yoga datasets using metrics such as Mean-squared deviation, cost over features, number of drifts detected, and accuracy, demonstrated to be faster and more accurate than IKS, WRS, an static classifier for baseline, and dynamic classifier with persistent detector for top line.

The Bayesian Nonparametric unsupervised framework, referred by BNDM \citep{ref111} utilizes Polya tree hypothesis test  decomposing the data distributions into a multi-resolution representation. It has been evaluated on Keystroke, Insects, Arabic, Bike, and Posture datasets using accuracy, and detection rate as their chosen metrics to compare with KS, RDa, RDm, MD3, PH, UE, HHT-u, baseline and topline classifiers, determining that this method is generally more sensitive to sudden and non-fixed space drifts comparing with other drift types.

A single windows method based on Multi-Armed Bandit, referred as f -dsw TS \citep{ref22}, enhancing the traditional Thompson Sampling approach along with the Discount factor, applying decay to past rewards, and giving more weight to recent observations. This method has been evaluated on The Baltimore Crime, Insects (Different variants), Local news, and  Air Microbes using Regret, Cumulative reward, and accuracy as their choice of metrics, and compared with other MAB methods, such as Thompson Sampling, Discounted TS, Sliding TS, min-dsw TS, mean-dsw TS, and max-dsw TS, where the proposed method emerged as the best performing MAB algorithm.

The one-class drift detector for unsupervised learning (OCDD) \citep{ref41} utilizes one window to monitor deviations that may indicate concept drift. This approach has been evaluated on a sum of 13 synthetic and real-world datasets and compared with other 17 methods using accuracy and determining that it outperforms the other methods by producing models with better predictive performance.

The MIS-ELM \citep{ref67} extends the ELM approach by applying a soft one-hot encoding on the categorical data, and operates incrementally by updating the model with incoming data to adapt to evolving patterns, and during stable periods, it leverages unlabelled data to fine-tune the existing classifier. This method has been evaluated on Adult, census, credit, NSL-KDD2, NSL-KDD5, mushroom, and Nursery datasets using accuracy and time overhead, then the results were compared with other ELM based methods, such as RELM, SS-ELM, US-ELM, ILR-ELM, OS-ELM, OR-ELM, WOS-ELM, and ESOS-ELM, where the results proved it improves the performance of ELM methods, while the computational cost showed a trade-off.

Another unsupervised method uses PCA, clustering using K-means, and Anderson-Darling statistical test to compare and detect significant differences \citep{ref105}, referred to as CSDDM. It has been evaluated on SEA, Hyperplanes, Elec2, and COVTYPE using metrics such as Accuracy, Number of drifts, and percentage of label used, TPR, FPR, ROC, and Execution time, where the results have been compared against an Static classifier, sliding window detector, and SCARGC, showing that CSDDM cannot detect the real concept drift as accurate as a supervised one, but with lower false alarm rate and competitive classification accuracy than other unsupervised approaches.

The CPSSDS method \citep{ref100} applies an incremental classifier with self-training to enhance learning from scarce labelled data by applying Conformal Prediction to identify unlabelled data points. It has been evaluated on SEA, STAGGER, AGRAWAL RBF, Waveform, Elect2, Spam and Vehicle datasets using accuracy as the choice of metric, and compared with Amanda, Fast Compose, STDS, Sco-Forest, SCBELS integrating HT and NB as base classifiers, where the CPSSDS showed it superiority against the other methods compared in the experiments.

The Type-LDA \citep{ref65} aims to identify the type of drift and the drift location point using a pre-trained drift type classifier on synthetic data, helping the base model to effectively select appropriate retraining strategies based on a sharing loss and maximising the Gaussian likelihood function. It has been evaluated on Toy dataset generated with MOA, Elec2, Weather, Spam, Airlines, and Poker hand using metric such as F1-score, Accuracy, running average time, and drift detection number. When compared with DWIN, DDM, EDDM, HDDM-A, HDDM-W, Page-Hinkley, KSWIN, AE, BLAST, and KUE, this approach proved that accurately identifying drift type and improve its adaptation accuracy better than the mentioned ones.

Another methods focused on time-series data employing Kolmogorov-Smirnov statistical distance, referred as WD \citep{ref79} to compare corresponding periods. It has been evaluated on 4 real datasets and 10 reproducible synthetic datasets using KS distance and accuracy as the metric chosen, and compared against five detectors, determining that it detects drift efficiently with minimal false alarms and has efficient computational resource usage. Similarly, the slidSHAPs method for multivariate time series \citep{ref11} that employs Shapley values to quantify the contribution of each variable to the overall correlation structure of the sliding windows. It has been evaluated on LED, ADD, MUL, COE, AND, OR, XOR, MIX, BC, PH, KDD, MSL datasets using Precision, recall, F1, and Average delay, and when compared to HDDDM, and ADWIN, it proved a dominating performance on almost all datasets.

\subsubsection{Ensemble-Based Drift Adaptation Methods}
In the other side, the methods studied for the ensemble-based type, we are able to find supervised approaches such as the NN-DVI \citep{ref68} that directly addresses the root causes of concept drift, consisting of three parts, where the first part follows the data modelling approach that retrieves the critical information from the dataset using a nearest neighbour-based partition schema, the second part is a distance function that accumulates regional density changes to quantify the overall discrepancy between the datasets, and the last component is a tailored statistical significance test for the distance. This approach has been evaluated on Elec2, Weather, and Spam datasets using accuracy, detected drifts, False detection rate, and False alarm rate, where it was compared with MMD, KL, CM, SAM-KNN, HDDM, ADW-Vol, AUE2, HAT, and ADWIN, showing that NN-DVI is capable of detect drifts accurately with similar performance. 

Another method based on multi-kernels with boundaries and centres of classes in the feature space \citep{ref94}, called MUKERS, where the data gets mapped into a Hilbert space in a search of linear relationships, and the multiple kernels actuate as different classifiers in an ensemble. This approach has been evaluated on SynCN, KDDCup99, and CoverType datasets using Accuracy, time running and memory as their metrics of choice, and compared with Loce, CLAM, ECSMiner and SvsClass, where its performance was similar in the accuracy level, with a very high computational complexity, causing delays. A combination of DDM or EDDM with LightGBM \citep{ref1}, proposed for cybersecurity purposes to detect and adapt to concept drift, where the data received is passed into the base learner, and its output is passed to the detector, which will classify the input as a warning, no drift level or drift level as soon as the data arrives. If no drift level, the algorithm continues to train from the data, if the warning level is raised, the data instances are stored, so when the drift level is recognised, a new model is trained using the same data instances that were stored during the warning level. This approach has been evaluated on Power grid data using metrics such as Precision, Recall, F1, and Accuracy, then compared to Adaboost, XGBoost, SAE DL, NNGE+STEM, and a CNN, where the results showed an increase on the accuracy, that reached 97.73\%. An improvement of the Adaptive Random Forest algorithm that implements the Resampling Effectiveness Measure \citep{ref3} provides an important contribution to the improvement of these algorithms combining accuracy and execution time to evaluate the efficiency of the methods. In addition, they also proposed an empirical solution to select the parameter for the Poison distribution used during the resampling process. This approach has been evaluated on LED, SEA, Agrawal, RTG, RBF, Hyperplane, Amazon customer review, Hotel Arabic-Reviews, and COVID-19 datasets using metrics such as Resampling effectiveness based on accuracy and execution time. It has been compared with ARF to determine that it exhibited considerable improvement than the base model. 

A mini-batching strategy \citep{ref20} aimed to increase the performance of bagging ensemble algorithms by aggregating incoming data in batches before it gets processed, improving memory access locality and reducing cache misses, helping to speed up to 5 times faster. This approach has been evaluated on Airlines, GMSC, Elec2, and Covertype datasets using metrics such as distance analysis, Memory footprint, cache usage, Accuracy, precision, and recall, where the results are compared with Lbag, ARF, SRP, OBAdwin, OBASHT, and OB proving it can improve the energy efficiency of bagging ensembles, and alleviate the memory bottleneck, increasing the speedup to some extent. 

The SEOA \citep{ref45} framework, enhances model convergence by integrating shallow and deep features through adaptive depth units, which dynamically manage information flow based on changes in adjacent data streams. Each adaptive depth unit functions as a base classifier within an ensemble, with weights adjusted in real time according to individual performance losses. Furthermore, SEOA employs a dynamic selection mechanism that responds to data fluctuations, balancing stability and adaptability. It has been evaluated on SEA, Hyperplane, RBF, LED, Tree, Elec2, KDDCup99, Covertype, and Weather using Average accuracy, accuracy, final cumulative accuracy, and recovery speed under accuracy (RSA), then compared with DNN-2, DNN-4, DNN-8, DNN-16, Resnet, Highway, and HBP, where it has proven that it can effectively contend with different types of concept drift and has good robustness and generalization. The SDDE method \citep{ref60} expands the previously proposed SDDM method, which employs an ensemble of statistical tests, and different measures that are calculated based on the Kernel Density Estimation to monitor different aspects of the data distribution that can lead to drifts. It has been evaluated on Elec2, Banknote, diabetes, and Wisconsin using Drift magnitude, conditioned marginal covariate drift, and Accuracy, then compared with DDM, EDDM, ADWIN, and HDDM, where the results obtained clearly outperformed the compared methods. The ACC\_KT approach \citep{ref43} extracts valuable knowledge from pre-drift data, transferring it to post-drift data by keeping historical model data that contains the key data to be passed to the ensemble classifiers, so every time a new one is trained, the new data is passed along with the key data. When drift occurs, a new model is trained, in case the total number of classifiers is lower than the maximum, a new model is then added, if the maximum is reached, the new model replaces the classifier that showed the lowest performance. It has been evaluated on Hyperplane, Sea, LED, RBFblips, Tree, Electricity, Weather, KDDcup99, and Covertype using Average accuracy, average AIF, and average recoverability, where the results compared with DNN-4, ResNet, Highway, HBP, DenseNet base classifiers with the methods DWM, SRP, DWCDS, ARF, SAL, and DTEL showed that this approach provides an acceleration effect, but still require more efficiency on different types of drifts and optimisation.

It is also possible to find unsupervised methods, such as the EDFS \citep{ref61} that recognise local changes in feature subspaces, they are created in a greedy manner using a round-robin strategy, and using specialised committees on incremental Kolmogorov-Smirnov tests, where a voting scheme is applied among all these tests, ensuring the creation of meaningful feature selections. It has been evaluated on Hyperplane, LED, RBF, SEA, and Tree datasets using Prequential accuracy as the metric chosen, and compared with RFS, lKS, LDCNet, and EHCDD, where the results shows that this approach offers significant benefits as compared with univariate approaches, where its performance can be compared with supervised methods. A framework that integrates drift detection and diagnosis \citep{ref25} is applied for heterogeneous industrial production processes, leveraging ensemble learning, novelty detection, and continuous learning. This method has been evaluated on private data using accuracy, precision, recall, F1-score, variance, Euclidean and Manhattan distances, and silhouette score, that compared with its different variations mixing KNN, SVM, and LSTM, where the results obtained are promising, but still requiring more optimisation to be effective. 

The DSE-DD approach \citep{ref24} which can be applied to any DNN, focused on high-dimensional image data. It uses multi-layer DNN activations along with the CBRI descriptor generation technique, as the activation reduction method, to identify concept drift and adapt to it. It has been evaluated on Fashio-MNIST, CIFAR10, and CIFAR100 using Average f1-score, Average accuracy, and standard deviation, then compared with DDM, ADWIN, HDDMW, KSWIN, MINAS, and OCDD, determining that it outperforms other methods by adding a 5-20\% accuracy increase, and reducing the adaptation time nearly 8 times less with 20.5s and 85\% accuracy post adaptation.

\subsubsection{Semi-Supervised and Hybrid Frameworks}
Between the semi-supervised approaches, a method based on a distributed Genetic Programming algorithm able to cope with multiple drifts \citep{ref36} that utilises a very efficient detection function based on fractal dimension along with a GP-based decision tree, that generates a new set of labelled tuples when drift is detected, and used it to train new classifiers that will be added to the ensemble, removing the oldest classifiers, so the overall number remains the same. This method has been evaluated on Sine, Hyperplane, Line, Plane, and Circle datasets using Accuracy, Error rate, miss detection rate, and false alarm rate, that after being compared with ADWIN, DDM and STEPD, it has been determined that it is comparable and also better of the other population-based approaches, but slower than RF approaches. The OPF classifier \citep{ref53} utilises a graph partition task, where each node is encoded by a sample that is connected to others using a predefined adjacency relation by measuring the path-cost using any smooth path-cost function. During the training, the algorithm builds the OPF that is later used in the training to conquer the specific nodes, providing the label of its conqueror with the updated cost. This method has been evaluated on Hyperplane, Usenet, SEA, Covertype, elec2, and Poker hand using Accuracy as the chosen metric to compare with other OPF versions, such as OPF-nomemory, OPF-fullmemory, and OPF-window3, showing that OPF ensemble is suitable compared to traditional OPF methods, where the ensemble showed higher accuracy.

When referring to Meta-learning solutions, the main goal is to produce models that have been trained in a wide number of learning tasks, so when they are utilised with new tasks, they can learn with fewer training samples. Kim and Jeon \citep{ref58} proposed an approach that uses the logic outputs from pre-trained networks and detects data drift by leveraging internal features extracted by vision models, such as ResNet and Vision Transformers (ViT) \citep{ref32}, instead of from the raw data. It has been evaluated on ImageNet-1k using Accuracy as the metric used to compare the results with other ResNet (50, 101, 152) and ViT (small, base, large), where the results reflect that this lightweight implementation can effectively identify drifts with high accuracy.

\subsubsection{Autoencoder and Representation-Learning Methods}
In addition, other different strategies and methods have been covered in this study, such as the use of autoencoders, small neural networks that are used to reconstruct input data and capable of learning compact nonlinear codes that summarise the most important features. An unsupervised approach \citep{ref55} demonstrates that by making use of these patterns from one part of the data stream, it can be compared in the incoming parts of the data stream, and monitoring for possible changes in the cost functions, such as the cross-entropy and the reconstruction error, identifying the drift and proceeding with the retraining of the model. It has been evaluated on synthetic datasets using Reconstruction error, and Cross-entropy then the results are compared with RBM being able to demonstrate that autoencoders can be used as detectors. Similarly, the AECDD approach \citep{ref57} employs the LSTM autoencoder, learning the parameters to represent the historical data, and during the detection phase, the reconstruction error is measured and compared with the historical using the Kolmogorov test, determining that the drift is happening. This approach has been evaluated on the SINE dataset using F1-score, accuracy, precision, and reconstruction error, where the results were compared with D3, HDDDM, and CDBD demonstrating it achieves better results even when operating with a non-optimal training size but needs to be extended to multiple drifts. 

Also, the VAE-CLT approach \citep{ref37} aims to identify abnormal behaviours in older adults by using data obtained from different sensors, implements a combination of Variational Autoencoders (VAEs) and the Central Limit Theorem (CLT), along with the Kullback-Leiber divergence between the daily activity map from sensors, and the baseline map generated from normal activity patterns. It has been evaluated on Synthetic dataset and data generated with sensors on 20 participant apartments using mean, variance, f1-score, Accuracy, Precision, and Recall, without being compared with other methods, their results are interesting, where they demonstrated to capture the behaviour of humans and detect changes by applying the drift detection methods to their patterns of movement.

\subsubsection{Semi-Supervised Neural and AutoML Frameworks}
A semi-supervised framework designed to detect concept drifts that adversely affect classification performance, known as real drifts, while ignoring those that do not and termed virtual drifts \citep{ref21}, this approach involves training a neural network to learn a low-dimensional embedding of input data, constrained to ensure that samples from the same class are closely clustered, and those from different classes are well-separated. During inference, the system monitors the distances between incoming data points and these class centroids in the embedding space. Significant deviations from established patterns signal a potential concept drift. This method has been evaluated on RBF and MovingRBF, Adult, bank, digits08, digits17, musk, phishing, and wine datasets using Detection accuracy, detection delay, Accuracy, and H-score, then compared with HDDDM, IKS, EMAD, ZSD versions with and without the proposed embedding framework, to determine that it effectively identifies drifts impacting the model's predictive accuracy without requiring access to ground truth labels during deployment. 

The NRSB \citep{ref44}  aims to reduce the negative effect of noise samples near category boundaries, introducing extension and intension distances to define the soft boundaries, and effectively segmenting the feature space. The method implements different labelling strategies depending on the new samples, where automatic labelling is applied to in-class regions, active expert labelling is applied in ambiguous mixed-class regions, and collaborative labelling for novel concepts. It has been evaluated on Hayes-Roth, Page blocks, Yeast, Poker hand, Forest, and KDDCrup99 using Novel precision rate, and Accuracy, then compared with AdBoostM2, Bag, TotalBoost, and MC\_SVMA to determine its superior performance, only some experiments identified MC\_SVMA as the best. An AutoML study on how to enhance these systems for evolving data \citep{ref23} proposes six adaptation strategies tailored to detect and respond to various types of concept drift, thereby maintaining model performance over time, and improving the resilience of AutoML systems to data evolution. They have been evaluated on Airlines, Elec2, IMDB, Vehicle, SEA, and Hyperplane datasets using Accuracy as the chosen metric to compare the results with Oza bagging and Blast, determining that these strategies proposed allow the autoML to detect, adapt, and recover from drifts successfully.

\scriptsize
\begin{longtable}{@{}>{\raggedright\arraybackslash}p{2cm}p{1.2cm}p{1.6cm}p{1.8cm}p{1cm}p{2.6cm}@{}}
\caption{Summary of Concept Drift Approaches\label{tab:concept_drift}}\\
\toprule
\textbf{Ref} & \textbf{Section} & \textbf{Category} & \textbf{Method} & \textbf{Drift} & \textbf{Datasets} \\
\midrule
\endfirsthead

\caption*{\textit{Table \thetable{} (continued)}}\\
\toprule
\textbf{Ref} & \textbf{Section} & \textbf{Category} & \textbf{Method} & \textbf{Drift} & \textbf{Datasets} \\
\midrule
\endhead

\bottomrule
\endfoot
{\citep{ref19}}  & Sliding Windows & Supervised & FPDD, FSDD, FTDD & Abrupt, gradual & 24 artificial datasets with LED, Mixed, RandomRBF, and Sine generators; Airlines, Covertype, Poker hand \\
{\citep{ref67}}  & Sliding Windows & Semi-supervised & Incremental Semi-supervised Extreme Learning Machine (MIS-ELM) & N/A & Adult, Census, Credit, NSL-KDD2, NSL-KDD5, Mushroom, Nursery \\
{\citep{ref100}} & Sliding Windows & Semi-supervised & Conformal prediction for semi-supervised classification (CPSDDS) & Various & SEA, STAGGER, AGRAWAL, RBF, Waveform, Elec2, Spam, Vehicle \\
{\citep{ref17}}  & Sliding Windows & Unsupervised & Least Squares Density-Difference (LSDD) & Abrupt & N/A \\
{\citep{ref75}}  & Sliding Windows & Unsupervised & Plover & N/A & Synthetic, S\&P500, Electricity, Airlines \\
{\citep{ref40}}  & Sliding Windows & Unsupervised & D3 & N/A & Electricity, Covertype, Poker hand, Rialto, Rotating Hyperplane, Moving RBF, Interchanging RBF \\
{\citep{ref92}}  & Sliding Windows & Unsupervised & Unsupervised Concept Drift Detection (UCDD) & Various & Synthetic \\
{\citep{ref96}}  & Sliding Windows & Unsupervised & Image-based Drift Detector (IBDD) & Various & Heartbeats, Flying insects, Sensor Posture, StarLightCurves, UWaveGestureLibrary, Yoga \\
{\citep{ref111}} & Sliding Windows & Unsupervised & Bayesian Nonparametric Drift Detector (BNDDM) & Sudden drifts & Keystroke, Insects, Arabic, Bike, Posture \\
{\citep{ref41}}  & Sliding Windows & Unsupervised & One-Class Drift Detector (OCDD) & N/A & 13 Synthetic and real-world datasets \\
{\citep{ref105}} & Sliding Windows & Unsupervised & Clustered Statistical Test for Drift Detection Method (CSDDM) & N/A & SEA, Hyperplanes, Elec2, Covertype \\
{\citep{ref10}}  & Sliding Windows & Unsupervised & Uncertainty Drift Detection (UDD) & N/A & Friedman, Mixed, Air Quality, Bike Sharing, IncAbr, IncReo, KDDCup99, Gas sensor, Elec2, Rialto \\
{\citep{ref11}}  & Sliding Windows & Unsupervised & SlidSHAPs & N/A & LED, ADD, MUL, COE, AND, OR, XOR, MIX, BC, PH, KDD, MSL \\
{\citep{ref35}}  & Sliding Windows & Unsupervised & Concept Drift Detection based on Jensen–Shannon Divergence & N/A & Artificial, Real-world \\
{\citep{ref116}} & Sliding Windows & N/A & Labelless Concept Drift Detection and Explanation (L-CODE) & N/A & SEA, Electricity, Phishing, Adult, Rabobank \\
{\citep{ref9}}   & Sliding Windows & N/A & Bhattacharyya Distance-based Concept Drift Detection Method (BDDM) & Gradual, abrupt & Agrawal, LED, RBF, Electricity, Shuttle, Connect-4, Diabetes, Airlines, Poker hand, Covertype, Pen digits, Gesture, Codrna \\
{\citep{ref22}}  & Sliding Windows & N/A & f-Discounted-Sliding-Window Thompson Sampling (f-dsw TS) & All & Baltimore Crime, Insects (variants), Local news, Air Microbes \\
{\citep{ref79}}  & Sliding Windows & N/A & WinDrift (WD) & N/A & 4 real and 10 reproducible synthetic datasets \\
{\citep{ref26}}  & Sliding Windows & N/A & Multi-level Weighted Drift Detection Method (MWDDM) & Abrupt, gradual & SINE, MIXED, CIRCLES, LED, Elec2, Forest Covertype, Poker hand \\

\end{longtable}

\subsubsection{Key Insights}

The approaches discussed in this section demonstrate the increasing role of probabilistic reasoning and ensemble based adaptation in handling concept drift. Bayesian windowing methods such as \textbf{SABeDM} provide explicit representations of uncertainty, which is particularly useful when observations are sparse or imbalanced. Multi level schemes such as \textbf{MWDDM} introduce structured warning and drift states that support smoother transitions between models. Ensemble techniques including \textbf{SEOA} and \textbf{OWA} reveal two complementary adaptation principles, namely the adjustment of predictive functions through selective weighting and the stabilization of parameters through temporal averaging. These developments indicate that modern concept drift handling integrates uncertainty management with both function space and parameter space adaptation in order to maintain reliable performance under evolving distributions.

\subsection{Catastrophic Forgetting}

This challenge refers to the phenomenon where a model or system abruptly loses previously acquired knowledge or skills when learning new tasks or adapting to changing environments. This typically occurs because the evolving process updates or replaces parts of the model, often optimizing for recent performance without preserving earlier adaptations. As a result, knowledge accumulated during previous evolving stages can rapidly degrade or disappear entirely, significantly impacting the model's ability to perform effectively on past tasks. This issue stems from the stability-plasticity trade-off: a model must be plastic enough to integrate new patterns but stable enough to prevent the overwriting of existing ones. 

A foundational review providing critical context for data-evolving scenarios \citep{ref82} introduced a taxonomy in the extended study on catastrophic forgetting for deep learning covering more than 200 approaches. Similarly, \citep{ref120} provides an extensive taxonomy, and an overview of different methods, determining the crucial stability-plasticity trade-off, and a comprehensive experimental comparison of 11 of them across multiple benchmarks. Another study focusing on Class-Incremental Learning (CIL) \citep{ref121} a scenario where the model must discriminate between all past and present classes without a task ID during inference. The paper provides a complete survey of existing CIL methods for image classification and conducts an extensive experimental evaluation on thirteen methods across various large-scale datasets and domain shift scenarios. Another article \citep{ref122} reviews studies that tackle catastrophic forgetting in deep learning models that rely on gradient descent, providing a new taxonomy, and identify gaps in the area.
\citep{ref123} systematically investigates different methods, focusing on deeply analysing their theoretical foundations, specific implementations, and their pros and cons with the aim of producing a classification of them,  where another review \citep{ref124} with the focus on visual tasks, where the authors propose a refined classification, and provides a deep analysis bases on their design principles, computational efficiency, and adaptability. Aiming to bridge the gap between basic settings, theoretical foundations, representative methods, and practical applications, the extensive article \citep{ref125} summarise methods with proper stabillity-plasticity trade-off, offering a detailed taxonomy and discussing promising future directions. A review that focuses specifically on the challenges of applying continual learning to Large Language Models (LLMs), where catastrophic forgetting remains a predominant problem after fine-tuning. It structures the field into two directions of continuity: vertical continuity (general to specific capabilities) and horizontal continuity (across time and domains). It reviews methods across Continual Pre-Training (CPT) and Continual Fine-Tuning (CFT) and highlights the need for new evaluation benchmarks for LLMs \citep{ref126}. On the other hand, another work argues that forgetting can be beneficial \citep{ref127}, for privacy and storage purposes, exploring the phenomenon in non-continual learning contexts.
Understanding "why" networks forget is a prerequisite for developing robust mitigation strategies. Recent theoretical research has moved beyond empirical observation to analyze the internal dynamics of neural networks, where \citep{ref128} proposed the "Forward Explanation" framework, attributing forgetting to the convergence of task representations and the interference of interleaved representations during training. An structural analysis \citep{ref129} revealed that deeper layers are significantly more susceptible to forgetting than early layers, and the severity of forgetting is directly correlated with the semantic similarity between tasks. In addition, \citep{ref130} demonstrated that multitask and sequential learning solutions are often connected by linear paths of low error, suggesting that if a model remains within these "stable regions," forgetting can be minimized. Most recently, the field has sought deeper theoretical grounding, with \citep{ref141} investigating the underlying dynamics of forgetting specifically within two-layer convolutional neural networks, offering mathematical insights into feature evolution and interference. Table 3 provides a summary.

\subsubsection{Regularization-based methods}

These techniques introduce constrains to the loss function to protect parameters critical for old tasks. 
Between the weight importance category, Elastic Weight Consolidation (EWC) solution proposes straining important parameters to stay closer to their original values by implementing a function that proportionally decreases their value based on their importance on performance from the previous task, and weighted by utilising different functions, along with a constraint to them that uses a quadratic penalty. An example that utilizes the Fisher Information Matrix to identify and protect critical weights \citep{ref131}.
Synaptic Intelligence (SI) methods are built on the theoretical idea of synaptic consolidation in neuroscience,   \citep{ref132} calculates the importance by tracking a synapse's contribution to the reduction of the loss function.
An analysis \citep{ref80} aimed to integrate developments in task space modelling and correlation analysis, by implementing Task2Vec methods to estimate the actual measures, where they introduced the notion of Total Complexity and Sequential Heterogeneity of task sequences, that it is combined with the correlation analysis by integrating Pearson correlation coefficients. This has been evaluated on MNIST and CIFAR datasets using Average Error rate, Total complexity, Sequential heterogeneity, and Pearson correlation, where the results were compared with SI, and VCL.
\subsubsection{Parameter Adaptation}
These adaptation methods analyses the weights and biases obtained during the inference constantly, adapting the parameters to mitigate the challenges when they are detected, without involving the alteration of the structural topology group where parameters are changed as a method to mitigate the challenge.

Some of the initial attempts to update parameters used as a solution for catastrophic forgetting by implementing a two-network architecture, where the new items are learned by the first network concurrently with internal pseudo-items generated from the second network, reflecting the structure of the previously learned network, where it was experienced the possibility of selectively slowing down the learning on the weights for new tasks that are important and protecting the important weights of the old previously learned tasks, kind of humans when they expend long periods learning and perfecting skills, where an study proposed two different approaches \citep{ref83}, a Speculative Backpropagation method (SB) that aims to readjust the weights by calculating derivates of the error to replicate back the past knowledge, and a Biased weight update with Activation history (AH), where they were able to preserve the important weights from a previous task and confirm that there are two types of activated neurons behaviours, (a) While training one specific task, the neurons activated in the past are more likely to be activated in the future, and the deactivated neurons are more likely to be deactivated. (b) The activated neurons differ from task to task. Only activated neurons affect the inference result and play an important role in that task. This has been evaluated on MNIST dataset using Accuracy as the metric to compare with Gradient descent, Joint training, EWC, and SI using a MLP as baseline, where the results show that continual learning ability is improved when using SB and AH. Incremental Moment Matching (IMM) \citep{ref133} aims to find a parameter space that satisfies the posterior distributions of both old and new tasks by matching their statistical moments.

On the other hand, an attempt to mimic the human brain for sequential learning using a multilayer model following Occam’s razor principle, consisting of interleaving past examples with the current training batch has been extended with the proposal of five tricks to mitigate the shortcomings that restrain Experience Replay \citep{ref18}, covering Independent Buffer Augmentation (IBA), Bias Control (BIC), Exponential Learning Rate Decay (ELRD), Balanced Reservoir Sampling (BRS), and the Loss-Aware Reservoir Sampling (LARS). They have been evaluated on CIFAR-10, CIFAR-100, and Fashion-MNIST using Average accuracy, and execution times as metrics to compare with SGD, Joint training, A-GEM, GEM, HAL, iCaRL, and ER, where the results displayed that ER along with the tricks outperforms more sophisticated approaches. Moving toward architectural constraints, \citep{ref138} introduced Hard Attention to the Task (HAT), utilizing learnable attention masks to preserve task-specific parameters without affecting future learning.

In general, methods focus on the forgetting nature of the models, where Soares and Minku \citep{ref95} proposed the Online Weight Averaging (OWA) approach based on stochastic weight averaging, implementing an ensemble of neural networks that work as a single model, and combining snapshots of different time steps and utilise those weights by averaging them and update the model. This method has been evaluated on Sine1, Sine2, Agrawal3, SEA1, SEA2, STAGGER1, STAGGER2, Elec2, NOAA, chess, keystroke, Luxemburg dataset, Ozone, Power supply, Sensors 29 and sensors 31 datasets using G-mean, and prequential accuracy as metrics to compare with Ozabag, Ozabag-adwin, DWM, CSB2, SRPC, KUE, CALMID, MICROFAL, and SWA, where the results displayed that OWA clearly outperforms SWA, and deliver similar or better predictive performance than other ensemble methods.

\subsubsection{Structure Adaptation}
The structure adaptation methods aim for the modification of the network topologies, which can be done by adding or removing neurons or layers based on the new data requirements.

The Growing Neural Networks solutions aim to increase the number of neurons or layers on the topology, allowing the incremental addition of neurons and connections between them, helping to adapt to changes in the distribution of the data.

As an example, Zhong et al. \citep{ref117} propose a dynamic online learning model called Dynamically evolving Deep Neural Network (DeDNN), capable of structural self-adjustment in response to non-stationary data streams. DeDNN evolves from a basic multilayer perceptron, adding neurons based on local prediction errors and adding hidden layers guided by changes detected using Jensen-Shannon divergence. An adaptive memory mechanism preserves both local and global samples to avoid catastrophic forgetting as the network structure dynamically expands. This method has been evaluated on Airfoil, WineQ, Rate1, ETTh1, CaliF, Bitcoin, Rate2, AirQ, CondM, SP500, OilF, Rate and HouseEP datasets using RMSE, and Total time running as the metrics of choice to compare with NADINE, where the results showed clear superiority.

The dynamic NAS techniques automate the design of neural networks by enabling the models to grow, shrink, or modify their architecture dynamically during running time, and without forgetting previous knowledge, such as the DEN method \citep{ref113} uses an efficient way of using group sparse regularisation to dynamically decide the number of neurons to add at which layer, avoiding fully retraining the model, this is done by checking the loss is below a certain threshold, otherwise, it will expand its capacity and the hidden units that are deemed unnecessary from the training will be dropped altogether. The evaluation on MNIST-Variation, CIFAR-100, and AWA using Accuracy, AUGROC, capacity, and expansion performance, where the results have been compared with DNN-MTL, DNN, DNN-L2, DNN-EWC, and DNN-Progressive, helping to determine that DEN significantly outperform the existing methods. This is extended in the ADL approach \citep{ref7} with a flexible structure, that can be built from scratch using a construct, and utilises the Network Significance formula to drive the hidden nodes growing and pruning mechanisms. This method has been evaluated on MNIST, Weather, KDDCup99, SEA, Hyperplane, SUSY, Hepmass, RLCPS, and RFID localization data using classification rate, execution time (ET), HL, HN, and the number of parameters (NoP) as the metrics to compare with pE+, pE, DFN, and DNN, showing that it consistently outperforms the compared methods. Contrary to these methods, EXPANSE \citep{ref52} proposed expanding pre-trained deep learning models vertically, adding neurons to the existing layers rather than horizontally as other approaches suggest. It has been evaluated on MNIST dataset using Accuracy, and compared to different version of DNN, where the results demonstrated the effectiveness of using two-step training, and increasing the whole model’s learning capacity through vertical expansion of the model. The AutoGrow \citep{ref107}, a method that automates the depth discovery on Deep Neural Networks, beginning with a shallow seed structure, it adds new layers if it will improve the accuracy of the model until it cannot be further improved. This method has been evaluated on CIFAR10, CIFAR100, SVHN, FashionMNIST, MNIST, and ImageNet using metrics such as Accuracy, difference between accuracies in autogrow after and before training, and growing time, where the results were compared with different implementation of ResNet to conclude that this rapid growing outperforms more intuitively correct growing methods. Similarly, the introduction of Progressive Neural Networks \citep{ref88}, addresses the catastrophic forgetting challenge presented in Neural Networks by expanding the network with new neurons and lateral connections in charge of the transfer learning from older tasks, without modifying existing parameters. Between the current limitations of this kind of method, the growth occurs indefinitely with each new task, requiring manual intervention to include a new column, where a future optimisation will be required to provide the model with capabilities to avoid the creation of expansions while new tasks are like previous ones. This approach has been evaluated on Sample data from different task domains related games using ROC AUC, and transfer score metrics to compare with a baseline Reinforcement Learning model, where the results displayed that this method is immune to forgetting. The T-SaS method \citep{ref86}, is based on a Bayesian framework with a discrete distribution-modelling variable that helps to capture sudden distribution changes and adapt dynamically to them by selectively activating subsets of neurons. This method has been evaluated on Bouncing ball, 3 mode system, Dancing bees, elec2, and Traffic datasets using Accuracy, Normalized Mutual Information, and Adjusted Rand Index as metrics to compare with MOCA, AdaRNN, ARIMA, DeepAR, ARSGLS, KVAE-RB, and JT, where the results displayed a superior performance in detecting and adapting. 

\subsubsection{Replay and Memory-based}
Replay methods address catastrophic forgetting by mimicking biological memory consolidation, specifically the mechanism of revisiting previous experiences while learning new tasks. These approaches typically maintain a small buffer of data from prior tasks to regularize the training of the current task.

Early approaches focused on using this stored data to constrain optimization. For instance, Gradient Episodic Memory (GEM) \citep{ref134} utilizes a small subset of episodic memory to guide gradient updates. Rather than simply mixing old and new data, GEM treats the stored samples as inequality constraints, ensuring that updates for the new task do not increase the loss on previously learned tasks. \citep{ref137} established a seminal baseline with iCaRL, proposing an incremental classifier and representation learning framework that combines nearest-neighbor classification with exemplar-based rehearsal and knowledge distillation; iCaRL selects and stores a small subset of "exemplar" images that best represent the class mean for previous tasks. During training, it combines these stored images with new data and uses a distillation loss to ensure the network's predictions on old classes do not drift significantly.

Building on the concept of gradient interaction, Meta-Experience Replay (MER) \citep{ref135} integrates meta-learning into the replay framework. While GEM focuses on preventing interference (constraints), MER explicitly aims for gradient alignment. It optimizes parameters such that gradients from new and old tasks are aligned, thereby maximizing knowledge transfer while simultaneously minimizing interference.

However, the use of replay buffers can introduce unintended artifacts in large-scale incremental learning, such as "recency bias"—a tendency for the model to favor new classes over old ones. To mitigate this, \citep{ref136} introduced Bias Correction (BiC). This method applies a dedicated bias correction layer to rescale the output logits, effectively balancing the classifier's predictions between old and new classes to ensure fair evaluation across all learned tasks.

Recent work has expanded these paradigms into biological and non-Euclidean domains; \citep{ref139} demonstrated the efficacy of sleep-like unsupervised replay, mimicking biological memory consolidation to reduce forgetting.

The work presented in the study \citep{ref140} is not a method per se but a foundational study establishing that Graph Neural Networks (GNNs) suffer heavily from forgetting. It benchmarks existing strategies (like LwF and EWC) on graph topology data, finding that Replay is often the only effective strategy for graphs, while regularization methods often fail.

\scriptsize
\renewcommand{\arraystretch}{1.2} 
\begin{longtable}{@{}>{\raggedright\arraybackslash}p{1.5cm}p{2cm}p{3.2cm}p{3cm}@{}}
\caption{Summary of Catastrophic Forgetting Methods\label{tab:catastrophic_forgetting}}\\
\toprule
\textbf{Ref} & \textbf{Category} & \textbf{Method} & \textbf{Datasets} \\
\midrule
\endfirsthead

\caption*{\textit{Table \thetable{} (continued)}}\\
\toprule
\textbf{Ref} & \textbf{Category} & \textbf{Method} & \textbf{Datasets} \\
\midrule
\endhead

\bottomrule
\endfoot

{\citep{ref80}}  & N/A & Task2Vec & MNIST and CIFAR \\
{\citep{ref113}} & Supervised  & Dynamically Expandable Network (DEN) & MNIST-Variation, CIFAR-100, AWA \\
{\citep{ref7}}   & Supervised  & Autonomous Deep Learning (ADL) & MNIST, Weather, KDDCup99, SEA, Hyperplane, SUSY, Hepmass, RLCPS, RFID localization \\
{\citep{ref52}}  & Supervised  & EXPANSE & MNIST \\
{\citep{ref83}}  & N/A & Speculative Backpropagation (SB) and Activation History (AH) & MNIST \\
{\citep{ref18}}  & N/A & IBA, BIC, ELRD, BRS, and LARS & CIFAR-10, CIFAR-100, Fashion-MNIST \\
{\citep{ref95}}  & N/A & Online Weight Averaging (OWA) & Sine1, Sine2, Agrawal3, SEA1, SEA2, STAGGER1, STAGGER2, Elec2, NOAA, Chess, Keystroke, Luxemburg dataset, Ozone, Power supply, Sensors 29 and 31 \\
{\citep{ref117}} & N/A & Dynamically Evolving Deep Neural Network (DeDNN) & Airfoil, WineQ, Rate1, ETTh1, Calif, Bitcoin, Rate2, AirQ, CondM, SP500, OilF, Rate, HouseEP \\
{\citep{ref107}} & N/A & AutoGrow & CIFAR10, CIFAR100, SVHN, Fashion-MNIST, MNIST, ImageNet \\
{\citep{ref88}}  & N/A & Progressive Neural Networks & Samples from different task domains, related games \\
{\citep{ref86}}  & N/A & T-SaS & Bouncing ball, 3-mode system, Dancing bees, Elec2, Traffic \\
{\citep{ref131}}  & N/A & N/A & N/A \\
{\citep{ref132}}  & N/A & N/A & N/A \\
{\citep{ref138}}  & N/A & Hard Attention to the Task (HAT) & N/A \\
{\citep{ref133}}  & N/A & Incremental Moment Matching (IMM)  & N/A \\
{\citep{ref18}}  & N/A & Independent Buffer Augmentation (IBA), Bias Control (BIC), Exponential Learning Rate Decay (ELRD), Balanced Reservoir Sampling (BRS), and the Loss-Aware Reservoir Sampling (LARS & N/A \\
{\citep{ref134}}  & N/A & Gradient Episodic Memory (GEM) & N/A \\
{\citep{ref137}}  & N/A & iCaRL & N/A \\
{\citep{ref135}}  & N/A & Meta-Experience Replay (MER) & N/A \\
{\citep{ref136}}  & N/A & Bias Correction (BiC) & N/A \\
{\citep{ref139}}  & Unsupervised & Sleep-like & N/A \\
{\citep{ref140}}  & Unsupervised & N/A & N/A

\end{longtable}

\subsubsection{Key Insights}

The methods surveyed here reflect a growing emphasis on adaptive model capacity as a means of mitigating catastrophic forgetting. Approaches such as \textbf{DEN \citep{ref113}} adopt sparsity based strategies that encourage the reuse of existing representations while expanding capacity only when necessary, thereby supporting efficient and resource aware adaptation. This is all the more relevant when considering more efficient and greener architectures, as a higher proportion of frozen parameters directly reduces the amount of training required \citep{schwartz2020green}. Techniques such as \textbf{AutoGrow \citep{ref107}} illustrate the potential of depth expansion, although they also underscore the computational and latency constraints associated with increasingly deep architectures. These considerations are especially important in continuous stream settings in which task boundaries are not explicitly defined. The overarching insight is that model adaptation in EML requires mechanisms that balance stability and plasticity while preserving computational viability in real time environments.

\subsection{Skewed Learning}

This challenge refers to the difficulty of learning from data streams where the distribution of classes is highly imbalanced, and certain classes may become significantly underrepresented, leading to poor generalization and biased predictions.

Between the surveys visited for this challenge, Johnson and Khoshgoftaar \citep{ref56} surveyed 131 methods for deep learning with class imbalance. Untoro et al. \citep{ref102} provided a survey and comparison of 23 base classifiers, such as Decision Tree, K-NN, Naïve Bayes, and SVM methods on the UCI dataset, and Kumar et al. \citep{ref64} present a review of methods and their applications with a total of 25 research papers.

The resampling techniques aim to modify the distribution or size of data samples to improve performance in dynamic or imbalanced environments, where GABagging \citep{ref106} utilises a sample combination optimisation genetic algorithm that reduces part of the majority of the class samples, while maintaining the number of classes, creating new sets and combining them until a subset is reached with balanced samples, which then is used to train several classifiers. It has been evaluated on b1, breast, yeast, wisconsin, vehicle, segment, pima, haberman, glass, and spect datasets using TPR, and AUC, then compared with OverBagging, UnderBagging, ABBag, HSBagging and RBSBagging, where the results showed that this method has good prediction effect, but its time complexity is too high. Similarly, Aminian et al. \citep{ref4} introduced a technique that addresses imbalanced data in regression tasks, and specifically when rare and extreme values are of significance, employing sampling strategies, under-sampling and over-sampling, that are guided by Chebyshev’s. This method has been evaluated on 14 benchmark datasets using Accuracy, where the results proved its superiority. A hybrid approach \citep{ref29} processes incoming data in chunks, applying over-sampling to augment minority class instances and instance selection techniques to refine the dataset. This balanced data chunk is then used to train ensemble classifiers. The approach has been evaluated using benchmark datasets using F-measure, and G-mean, where the results show an improved performance of baseline classifiers.

On the other side, AMSCO \citep{ref66} implements two stochastic swarm heuristics optimizers, that cooperatively find a close to optimal mix with high accuracy and reliability. It has been evaluated on Abalone, Cleveland, Flare, glass, Haberman, Pima, Poker hand, vehicle, Wine, and yeast datasets using Accuracy, standard deviation, precision, recall, f1-score and Kappa as metrics, comparing the results with NN, Bagging, AdaBM1, CTS, DT, AdaC, SBO, RUS, BC, EE, and ADASY, where it was determined that it excellently overcomes other methods. An optimised Random-SMOTE \citep{ref114} that utilises K-means clustering to identify the centroids for the minority classes and calculate their distance to the centroids of the majority ones, where the samples with the smaller distance are then taken for oversampling using the Random-SMOTE algorithm to finish to balance the data in the data level. Then, the Biased-SVM algorithm is used to balance the data at the classifier level by setting the different penalty factors for the minority and majority samples. It has been evaluated on Banana, Haberman, Appendicitis, Vehicle, and Wisconsin datasets using Accuracy, and F1-score as chosen metrics to compare with SVM, and base Random-SMOTE, where it was determined that this approach adds some improvements on the classification effects, but the optimization of the parameters still requires more optimisation.

Under the cost-sensitive learning type of methods, the CMBoost \citep{ref115} algorithm is based on the cost-sensitive margin statistical characteristics, prioritising minority-class recognition, reducing overfitting, and obtaining fewer false negatives, that are combined with a boosting algorithm. This method has been evaluated on heart, sonar, vehicle, wine, wpbc segment, and vote datasets using Recall, precision, Positive Accuracy, Negative accuracy, F-measure, and G-mean as the metrics to compare with AdaBoost, where the results showed an increase in performance. On the other side, C-OSELM \citep{ref69} introduced an online sequential extreme learning machine, as the basic theory model, combined with a cost-sensitive strategy, to improve classification performance for the minority classes. This involves the assignation of different penalty parameters to the different classes, adjusting the misclassification costs and mitigating the decision boundary bias, by incorporating a cost adjustment function that dynamically optimises the weight parameters. This method has been evaluated on Pima, yeast, Haberman, vehicle, segment, yeast Ecoli, page-blocks, Vowel, Led, Shuttle, and Abalone datasets using G-mean as the metric used to compare the performance against OSELM and WOS-ELM, where the results displayed that it was significantly better in almost all the experiments. The CSRDA method \citep{ref27} extends the RDA approach by including a new optimisation function based on cost sensitivity that prioritises the correct classification of minority classes, and enforces sparsity when handling severely imbalanced datasets. This approach has been evaluated on six benchmark datasets using G-mean to compare with baseline methods, showing that it improves classification performance, and captures sparse features efficiently, resulting in better model interpretability.

Between the ensemble methods for this challenge, a hybrid approach that combines the SMOTE over-sampling technique, aiming to generate synthetic samples for the minority class, and the XGBoost algorithm, called SMOTEXGBoost \citep{ref73}, is well-known for its robustness. This method has been evaluated on German, Winsconsin, Glass, and E-coli using Accuracy, and AUC as metrics to compare with SMOTE, XGBoost, SMOTEXGBoost, LR, RF, and SVM, determining that even the results show an increase on the performance, it was proven the lack of stability of the approach. Another SMOTE over-sampling generation in combination with Gaussian Mixture Model \citep{ref101} where the synthetic data samples are selected based on the weight associated with each cluster and determined by the distribution of the majority class samples. This method has been evaluated on CM, PC, MC, MW, KC, and Haberman datasets using Precision, Recall, and sensitivity to compare with SMOTE, SMOTE-RUS, SMOTE-Tomek Links, and SMOTE-OSS, where the results concluded that it showed a higher sensitivity than others and performed better. WECOI approach \citep{ref30} integrates a weighted ensemble of one class classifiers with over-sampling and instance selection techniques to balance the class distribution, generating synthetic data samples for the minority classes. It has been evaluated on Heart, Diabetes, WBC, Acredit, Gcredit, Sonar, Satellite, Banana, Image, Thyroid, Spambase, Twonorm, Elec2, Airlines, Ozone, Gas, hyperplanes, AGR, and SEA datasets using Average accuracy to compare with AWE, HOT, iOVFDT, OUOB, OB, Learn++.NIE, and WECU, where the results demonstrated to be a competitive solution for them. Table 4 provides a summary.

\scriptsize
\renewcommand{\arraystretch}{1.2} 
\begin{longtable}{@{}>{\raggedright\arraybackslash}p{2cm}p{1.5cm}p{1.6cm}p{1.8cm}p{2.6cm}@{}}
\caption{Summary of Skewed Learning Methods\label{tab:skewed_learning}}\\
\toprule
\textbf{Ref} & \textbf{Section} & \textbf{Category} & \textbf{Method} & \textbf{Datasets} \\
\midrule
\endfirsthead

\caption*{\textit{Table \thetable{} (continued)}}\\
\toprule
\textbf{Ref} & \textbf{Section} & \textbf{Category} & \textbf{Method} & \textbf{Datasets} \\
\midrule
\endhead

\bottomrule
\endfoot

{\citep{ref66}}  & Resampling & Supervised & Adaptive Multi-objective Swarm Crossover Optimization (AMSCO) & Abalone, Cleveland, Flare, Glass, Haberman, Pima, Poker hand, Vehicle, Wine, Yeast \\
{\citep{ref106}} & Resampling & Supervised & GABagging & B1, Breast, Yeast, Wisconsin, Vehicle, Segment, Pima, Haberman, Glass, Spect \\
{\citep{ref114}} & Resampling & Supervised & Optimised Random-SMOTE with Biased-SVM & Banana, Haberman, Appendicitis, Vehicle, Wisconsin \\
{\citep{ref4}}   & Resampling & Supervised & Under and over sampling using Chebyshev’s & 14 Benchmark datasets \\
{\citep{ref29}}  & Resampling & Supervised & Hybrid oversampling and instance selection & Benchmark datasets \\
{\citep{ref115}} & Cost-sensitive & Supervised & CMBoost & Heart, Sonar, Vehicle, Wine, WPBC, Segment, Vote \\
{\citep{ref69}}  & Cost-sensitive & Not specified & C-OSELM & Pima, Yeast, Haberman, Vehicle, Segment, Ecoli, Page-blocks, Vowel, Led, Shuttle, Abalone \\
{\citep{ref27}}  & Cost-sensitive & Not specified & Cost-Sensitive Regularized Dual Averaging (CSRDA) & 6 Benchmark datasets \\
{\citep{ref73}}  & Ensemble & Supervised & SMOTEXGBoost & German, Wisconsin, Glass, E-coli \\
{\citep{ref101}} & Ensemble & Supervised & Cluster-based approach combining SMOTE oversampling with Gaussian Mixture Model clustering & CM, PC, MC, MW, KC, Haberman \\
{\citep{ref30}}  & Ensemble & Supervised & Weighted Ensemble with One-class Classification and Oversampling and Instance Selection (WECOI) & Heart, Diabetes, WBC, Acredit, Gcredit, Sonar, Satellite, Banana, Image, Thyroid, Spambase, Twonorm, Elec2, Airlines, Ozone, Gas, Hyperplanes, AGR, SEA \\

\end{longtable}

\subsubsection{Key Insights}

The studies reviewed in this section show that class imbalance interacts strongly with distributional drift and can significantly alter model behavior. Oversampling techniques such as \textbf{SMOTE} may become unreliable when the minority class shifts over time, since they can amplify outdated regions of the feature space and thereby degrade precision and recall. In contrast, approaches such as \textbf{C-OSELM} offer more stable performance by adjusting decision boundaries without synthesizing new samples, which reduces the risk of reinforcing obsolete structures. These observations emphasize that imbalance handling in EML must be designed with explicit consideration of drift dynamics to ensure robust model adaptation.

\subsection{Synthesis of Cross-Cutting Insights
}

Across the methods reviewed in this survey, a coherent view of EML emerges in which the tasks of detecting, interpreting, and adapting to change form an integrated pipeline rather than a sequence of isolated operations. In the context of data drift, approaches such as \textbf{DetectA}, \textbf{LASSO}, and \textbf{Adaptive SPLL} illustrate the progression from reactive error driven detection toward strategies that monitor input distributions, structural feature importance, and adaptive windowing. These developments emphasise the importance of balancing early sensitivity to change with robustness to noise and gradual evolution. 

Within concept drift and ensemble based adaptation, methods such as \textbf{SABeDM}, \textbf{MWDDM}, \textbf{SEOA}, and \textbf{OWA} demonstrate the growing role of uncertainty modelling, structured warning mechanisms, and both function space and parameter space adjustment. These techniques collectively highlight that modern drift handling must couple reliable detection with principled mechanisms for maintaining stability during transition.

In addressing catastrophic forgetting, approaches including \textbf{DEN} and \textbf{AutoGrow} reveal an increasing emphasis on adaptive model capacity. These methods encourage selective retention of prior knowledge and introduce new representational resources only when required, while also drawing attention to the computational constraints implicit in stream based and latency sensitive environments.

Finally, the analysis of skewed learning shows that imbalance handling interacts fundamentally with drift. Techniques such as \textbf{SMOTE} may reinforce outdated minority structures when class distributions move, whereas cost sensitive strategies such as \textbf{C OSELM} provide more reliable adaptation by avoiding assumptions about the location of minority examples.

Taken together, these insights indicate that EML is evolving into a unified framework in which systems must integrate proactive and uncertainty aware detection, informed interpretation of structural change, and adaptive mechanisms that span functions, parameters, and model capacity. The field increasingly recognises that effective learning under non stationary conditions requires coordinated design across these components rather than reliance on any single methodological perspective.

\section{Comparative Analysis and case studies}

This section presents a comprehensive comparative analysis of existing Evolutionary Machine Learning (EML) solutions and their suitability for real-world applications. It reviews both survey and research papers to identify key trends, benchmark methodologies, and evaluate the maturity of research within the field.

\subsection{Comparative Study of Methods}

\begin{figure}[h!]
    \centering
    \includegraphics[width=0.5\textwidth]{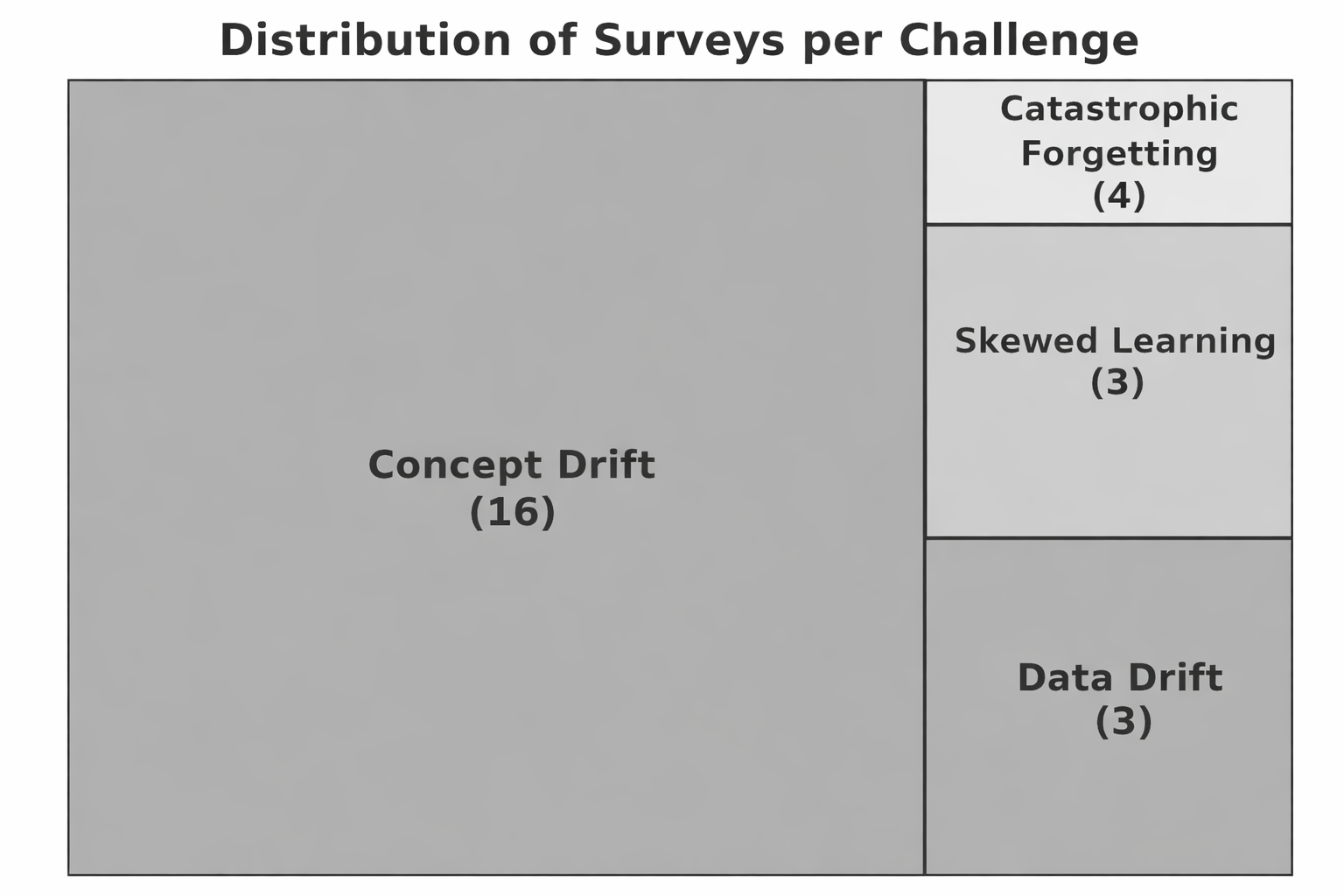}
    \caption{Tree map with Survey distribution per challenge}
    \label{fig:survey_distribution}
\end{figure}

A total of 26 survey papers were reviewed to capture recent developments addressing the primary challenges in EML.
Figure \ref{fig:survey_distribution} illustrates the distribution of these surveys across the identified challenges. The results indicate that Concept Drift is the most extensively studied area, encompassing 16 of the reviewed survey, followed by Catastrophic forgetting with 4 surveys, then Skewed Learning and Data Drift each represented by three surveys. Notably, a single survey covers all the related challenges comprehensively.

\begin{figure}[h!]
    \centering
    \includegraphics[width=0.5\textwidth]{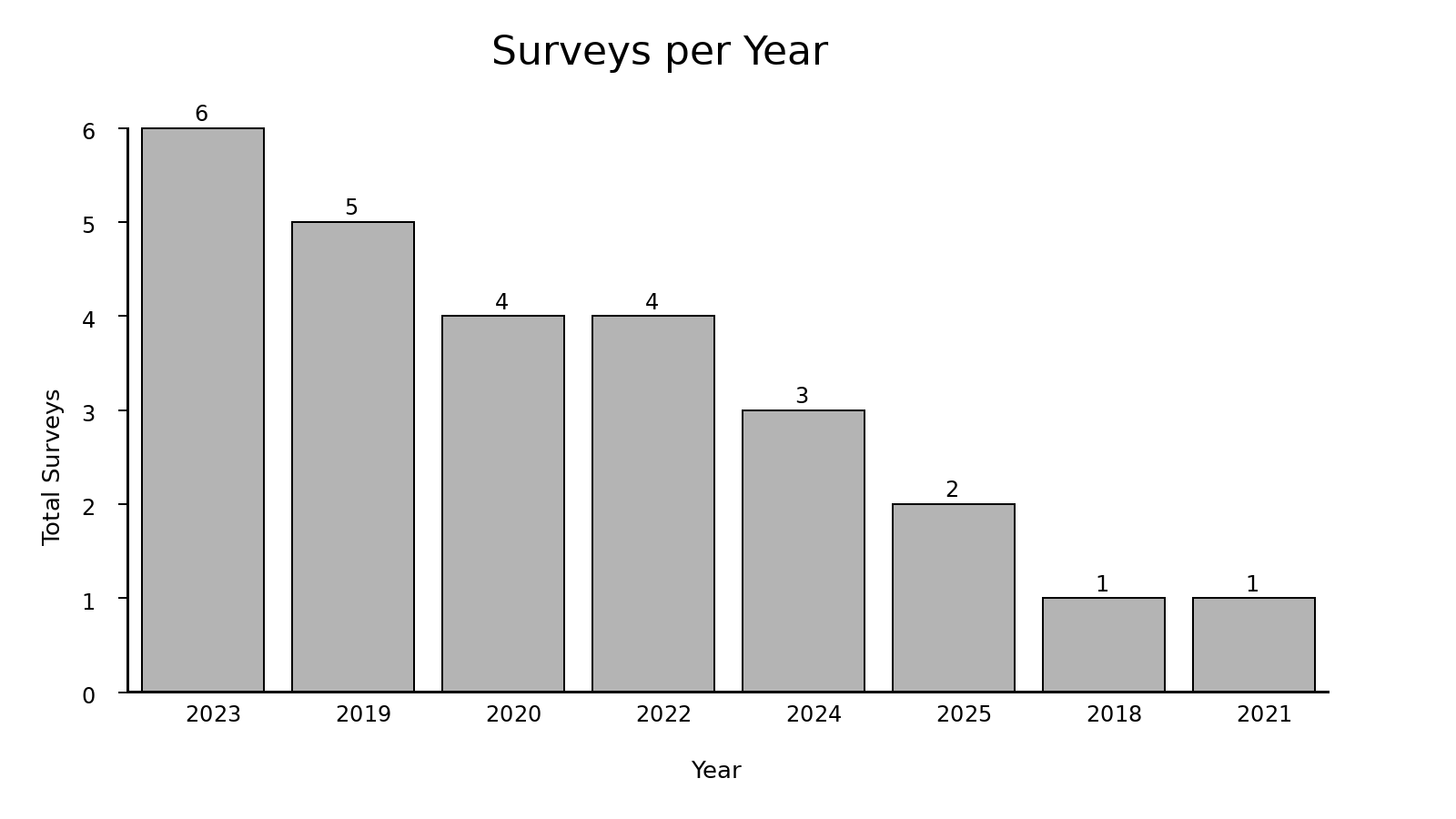}
    \caption{Bar chart with Survey distribution by year}
    \label{fig:survey_year}
\end{figure}

Figure \ref{fig:survey_year} presents the temporal distribution of survey publications. The majority of surveys reviewed were published in 2023, with six papers, followed by the year 2019 with 5 papers. Four were published in 2020 and 2022, three in 2024, and only isolated studies appeared in 2018 and 2021. This consistent publication activity demonstrates that EML remains an active and evolving research area.

\begin{figure}[h!]
    \centering
    \includegraphics[width=0.5\textwidth]{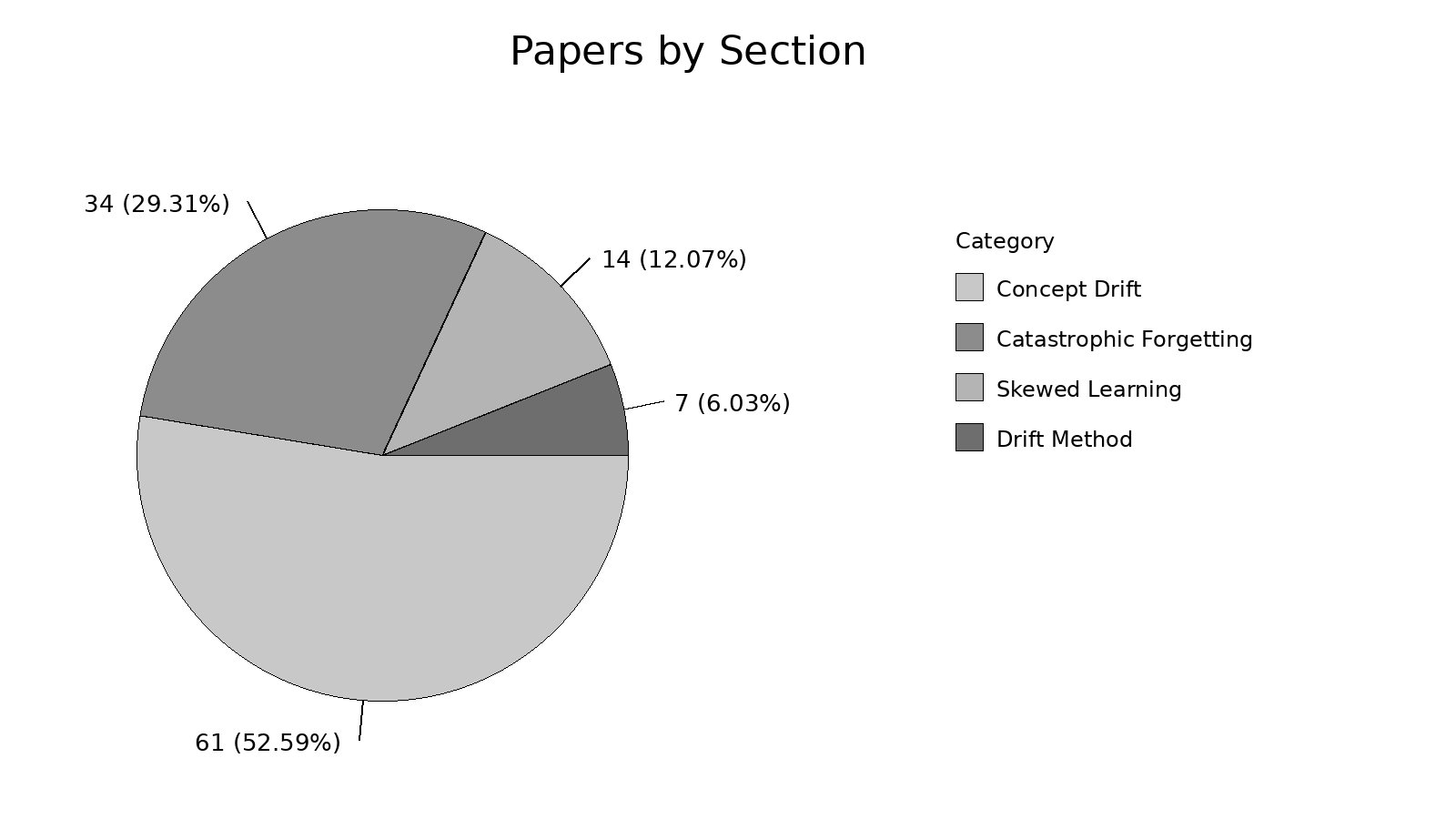}
    \caption{Pie Chart with reference distributions by challenge}
    \label{fig:ref_challenge}
\end{figure}

Beyond survey papers, this work also reviews 106 primary studies addressing the four key EML challenges. Figure \ref{fig:ref_challenge} depicts their distribution:
\begin{itemize}
    \item Concept Drift dominates with 61 studies,
    \item Catastrophic Forgetting with 34.
    \item Skewed Learning follows with 14,
    \item Data Drift with 7,

\end{itemize}

This uneven distribution reflects the maturity of certain subfields. Concept drift and catastrophic forgetting have been widely explored for over a decade. The “Others” category includes emerging topics such as federated EML and neurosymbolic adaptation, which, though limited in number, indicate promising research frontiers. The imbalance underscores both the established importance of drift detection and the growing opportunities in less mature subdomains.

\subsubsection{Data Drift}
\begin{figure}[h!]
    \centering
    \includegraphics[width=0.5\textwidth]{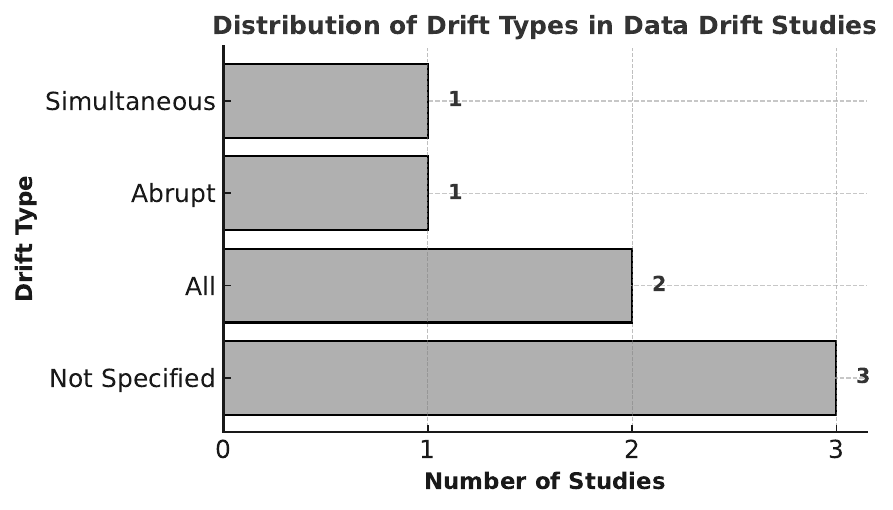}
    \caption{Bar chart with drift type distribution on data drift approaches}
    \label{fig:drift_types_distr}
\end{figure}

Analysis of the 10 data drift studies, see Figure \ref{fig:drift_types_distr}, shows that two methods addressed all drift types, while one each focused on abrupt and simultaneous drifts; three did not specify a drift type. Benchmark datasets include synthetic sources (Sine, Circles, Stagger, Hyperplane, SEA) and real-world datasets (Elec2, CoverType, Poker Hand, Rialto, Airlines, CIFAR10). Baseline comparison methods primarily involve DDM and EDDM, evaluated with metrics such as magnitude, accuracy, and prequential accuracy.

\subsubsection{Concept Drift}
\begin{figure}[h!]
    \centering
    \includegraphics[width=0.5\textwidth]{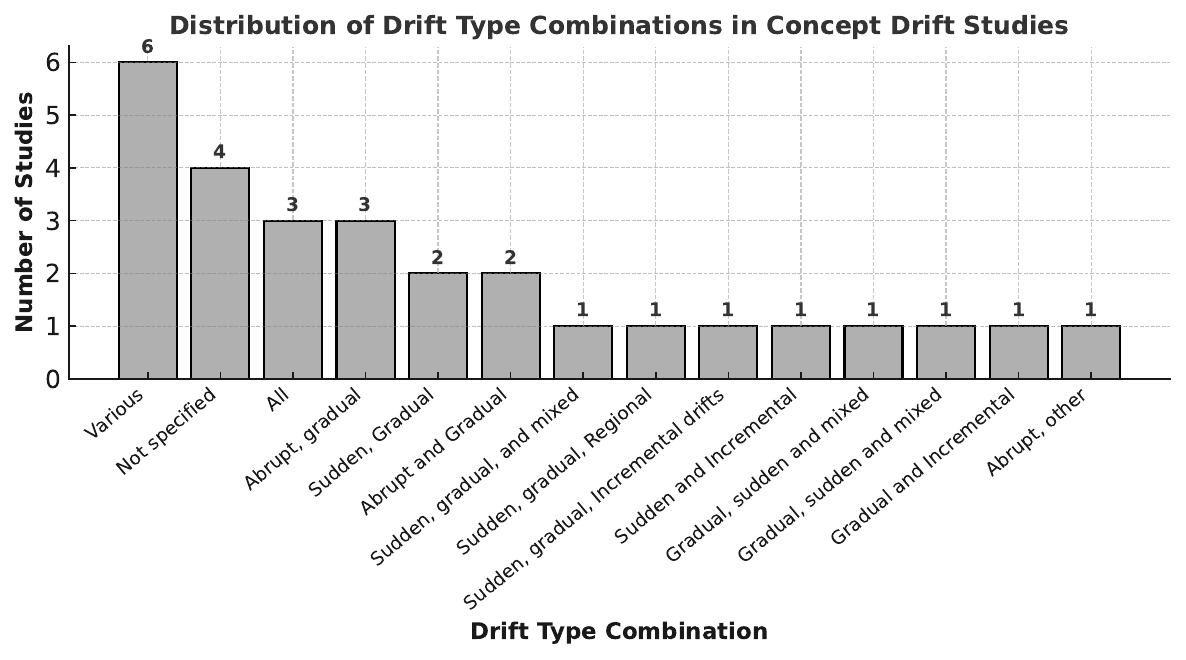}
    \caption{Bar chart the distribution based on drift type}
    \label{fig:drift_types_bar}
\end{figure}

Among the 29 concept drift studies, Figure \ref{fig:drift_types_bar}, a variety of drift types were reported. The “various” category is most common (six studies), followed by “not specified” (four), “all drifts” (three), and several combinations of abrupt and gradual drifts.

\begin{figure}[h!]
    \centering
    \includegraphics[width=0.5\textwidth]{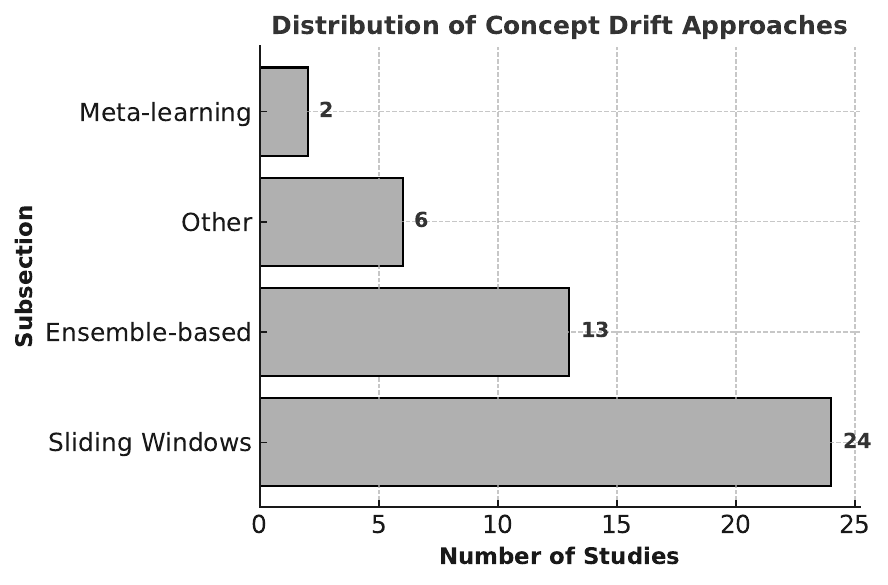}
    \caption{Bar chart with distribution based on the type of approach}
    \label{fig:drift_types_type}
\end{figure}

As shown in Figure \ref{fig:drift_types_type}, sliding-window-based methods dominate this area (24 studies), followed by ensemble-based (13), other (6), and meta-learning approaches (2).
Benchmarks often use synthetic datasets (Sine, Agrawal, SEA, Stagger, Hyperplane) and real-world datasets (Airlines, Elec2, MNIST, CIFAR, CoverType, Poker Hand, Rialto, KDDCup99). Common baselines include DDM, EDDM, ADWIN, HDDMW, SMOTE, STEPD, D3, and NEAT, with evaluation metrics such as accuracy, F1-score, G-mean, detection time, and number of drifts.

\subsubsection{Catastrophic forgetting}

\begin{figure}[!htb]
    \centering
    \includegraphics[width=0.5\textwidth]{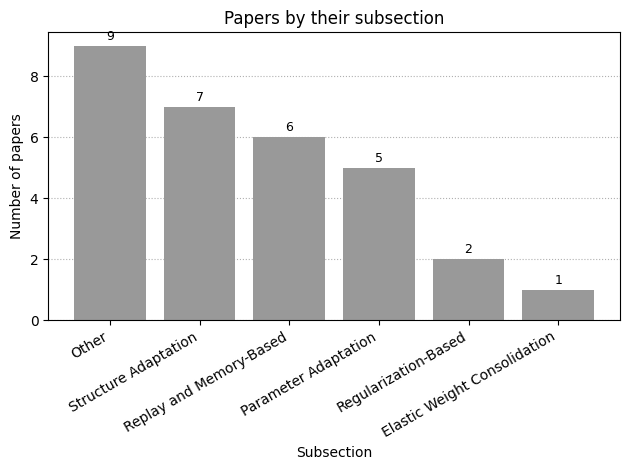}
    \caption{Bar chart with the distribution of approaches based on the section}
    \label{fig:distr_sec}
\end{figure}

The Catastrophic Forgetting challenge has been addressed by 30 main studies, see Figure \ref{fig:distr_sec}. Overall, the graph indicates that papers are concentrated in broader or structural adaptation categories, while more specific regularization-based methods, particularly Elastic Weight Consolidation, are comparatively rare. The most frequent approaches are those refereed as others with nine papers, that contains theoretical, empirical and different analysis, followed by structure adaptations with 7 papers, Replay and memory-based methods with 6 papers, Parameter adaptation includes 5 papers, Regularization-based are less common with 2 papers, and the least represented is Elastic Weight Consolidation with 1 paper. Benchmark datasets are primarily synthetic (SEA, Hyperplane) or standard real-world image datasets (MNIST, CIFAR variants). Unlike other sections, no common baseline exists, as each study employs specialized deep neural network (DNN) comparison models, evaluated primarily by accuracy and error rate.

\subsubsection{Skewed Learning}
\begin{figure}[h!]
    \centering
    \includegraphics[width=0.5\textwidth]{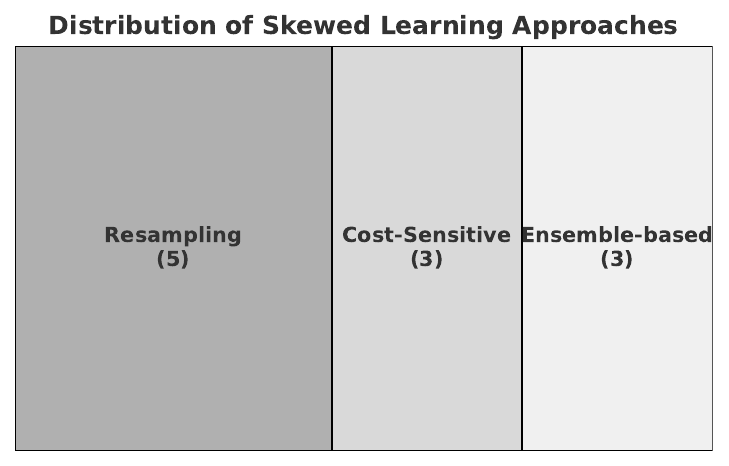}
    \caption{Tree map with section distribution for Skewed learning approaches}
    \label{fig:skew}
\end{figure}

The Skewed Learning category includes 11 studies, Figure \ref{fig:skew}. Most employ resampling techniques (five studies), followed by cost-sensitive and ensemble methods (three each). Common datasets include Yeast, Wisconsin, Vehicle, E-coli, Sonar, and Glass. Baseline models frequently use bagging and SMOTE variants, with performance measured via accuracy, G-mean, and F-measure.

\subsection{Suitability in Real-World Applications}

EML techniques have demonstrated applicability across a broad spectrum of domains. In image classification, evolutionary algorithms evolve feature extractors and neural architectures for higher efficiency and accuracy. In time-series prediction and anomaly detection, evolutionary strategies optimize parameters and ensemble structures, enhancing robustness to non-stationary data.
Applications also extend to medical diagnosis, spam detection, cybersecurity, and financial forecasting, all of which require adaptability under uncertainty.

Concept drift appears as the most represented domain across the studies, while other fields such as image classification, medical diagnosis, and case-based reasoning (CBR) are found. This diversity highlights the flexibility of EML in addressing visual recognition, sequential modeling, and dynamic decision-making problems.

Emerging application areas include activity recognition, cybersecurity, credit scoring, spam detection, and predictive maintenance, where EML improves adaptability and model generalization. Notably, in cybersecurity, evolutionary methods enhance resistance to adversarial attacks, while in medical domains, they optimize diagnostic models for better sensitivity and specificity.
Less conventional but promising applications include behavior analysis, microprocessor inspection, Othello gameplay, and gradient-based optimization, emphasizing EML’s potential for innovation in complex or evolving environments.

Several studies also explore socially significant use cases such as crime prediction, malaria vector classification, and attack detection, underscoring EML’s potential in public safety and epidemiology. These problems often involve heterogeneous data sources (e.g., sensors, text, and social networks), where evolutionary models effectively integrate multimodal signals.
Furthermore, the use of abstract problem categories (e.g., CBR, online learning, regression, XOR problems) illustrates EML’s expanding role in advancing learning paradigms rather than addressing isolated tasks.

Finally, computational efficiency varies considerably among EML methods. Lightweight approaches such as LASSO and OWA (O(n) complexity) are well-suited for real-time or edge applications, whereas ensemble and dynamic expansion methods (e.g., SEOA, DeDNN) are computationally intensive, requiring high-performance infrastructure. Balancing accuracy and latency is particularly critical for time-sensitive applications such as autonomous systems, where delays under 100 ms are often required.

\subsection{Case studies}

Several case studies demonstrate the practical implementation of EML in real-world environments:
\begin{itemize}
    \item Finance: “SEOA \citep{ref45} was deployed in a real-time stock trading system to adapt to market volatility on the NYSE dataset. The method achieved an 8
    \item Healthcare: “C-OSELM \citep{ref69} was applied to ICU patient monitoring using the MIMIC-III dataset. The model improved G-mean by 12
    \item Cybersecurity: “DetectA \citep{ref34} was utilized for network intrusion detection on the NSL-KDD dataset, identifying abrupt attack patterns with a 15
\end{itemize}

\section{Future Research Directions}

EML is still at an early stage, and several promising avenues require sustained investigation. Our survey identifies a set of future directions that build directly on the limitations of current methods and the gaps highlighted in the comparative analysis.

\textbf{Explainability in Adaptive Systems: }
While explainable AI is an established research field, little work addresses how explanations should evolve alongside the model itself. Current approaches to explainability assume static models, which limits their applicability in dynamic contexts where decision boundaries change over time. Future research must explore temporal explainability, allowing practitioners to track how predictions and model structures evolve under data drift or catastrophic forgetting. Such methods would not only improve trust but also support regulatory compliance, as the EU AI Act emphasizes traceability in adaptive systems. A key challenge is balancing the trade-off between interpretability and adaptation speed in real-time applications.

\textbf{Multimodal and Cross-Domain EML: }
Most evolving systems to date have been developed for unimodal data streams, often tabular or image-based. Yet many critical applications such as autonomous driving, healthcare monitoring, or industrial IoT generate multimodal streams where visual, textual, and sensor inputs evolve at different rates. Research is needed to develop adaptive fusion strategies that align drifts across modalities without destabilizing the overall model. Cross-domain adaptability also remains underexplored: models that can evolve simultaneously across related but distinct domains (e.g., adapting fraud detection models trained in one country to another with minimal retraining) would have significant practical value.

\textbf{Benchmarks and Evaluation Protocols: }
Our comparative analysis highlights inconsistent evaluation practices: studies use different datasets, metrics, and drift scenarios, limiting comparability. Establishing standardized evolving benchmarks covering synthetic, real-world, and multimodal streams would enable fairer comparisons. Beyond accuracy, benchmarks should evaluate adaptation delay, resource efficiency, fairness, and robustness. Platforms like OpenML could host evolving datasets with community-agreed protocols, ensuring reproducibility. Without such standards, progress risks becoming fragmented and difficult to measure.

\textbf{Efficiency and Resource Awareness: }
Many EML algorithms, especially deep evolving models, are computationally expensive, hindering deployment on edge devices such as wearables or mobile robots. Future work should focus on resource-aware EML, explicitly optimizing latency, memory, and energy consumption alongside accuracy. Promising directions include lightweight drift detectors, sparsity-inducing regularizers, and neuromorphic or FPGA-friendly architectures. Efficiency will be especially crucial in real-time safety-critical environments like autonomous vehicles or medical monitoring, where high latency may undermine system reliability.

\textbf{Human-in-the-Loop EML: }
Fully autonomous adaptation may not always be acceptable, particularly in safety-critical domains. Future work should therefore investigate human-in-the-loop frameworks where domain experts supervise, validate, or guide adaptation. Such approaches could mitigate risks of erroneous drift detection or biased updates. A research challenge is designing interaction protocols that allow human oversight without creating excessive cognitive burden or delaying adaptation in real-time applications.

\textbf{Ethical and Societal Dimensions: }
As EML systems become more pervasive, ethical issues move to the forefront. Adaptation itself may introduce or amplify bias if minority groups are underrepresented in evolving data streams. Future research should embed fairness-aware mechanisms within adaptation pipelines, for example through bias-sensitive drift detectors or regularizers. Transparency will also be essential, as evolving systems must remain auditable even as they change over time. Beyond technical solutions, collaboration with policymakers will be required to align EML with emerging governance frameworks.

Taken together, these directions indicate that the next wave of EML research should not only pursue algorithmic advances, but also tackle system-level integration balancing adaptability, interpretability, efficiency, fairness, and human oversight. Only by addressing these intertwined challenges can EML transition from an emerging research niche into a trusted paradigm for dynamic, real-world AI.

\section{Conclusion}

This survey analysed extensively influential works on evolving machine learning methods and techniques, specifically targeting critical challenges: data drift, concept drift, catastrophic forgetting, and skewed data. The extensive coverage includes 26 detailed surveys spanning all challenges and providing a foundational perspective on existing methodologies. Data drift, addressed through 7 key studies, highlighted innovative drift detection methods, adaptive data augmentation, and active learning approaches crucial for maintaining performance consistency in dynamically evolving data environments. The extensive exploration of concept drift with 61 studies emphasized effective strategies such as sliding windows for timely drift detection, ensemble-based adaptations for robust predictions, meta-learning methods facilitating rapid adaptation, and prototype-based learning approaches to manage evolving data distributions effectively.

The critical issue of catastrophic forgetting was carefully examined across 34 significant studies, detailing approaches like dynamic network expansion strategies promoting adaptable model architectures, and Elastic Weight Consolidation to promote a fairer distribution of weights for old values, helping them to maintain a grade of importance over the time. Meanwhile, skewed data challenges, discussed through 14 dedicated papers, underlined the efficacy of resampling techniques, cost-sensitive learning, ensemble methods, and fairness-aware models to address class imbalance and fairness constraints effectively.

Evaluation methodologies across these studies were diverse, incorporating a broad spectrum of synthetic and real-world datasets. Synthetic datasets allowed controlled simulation of specific drift conditions, catastrophic forgetting scenarios, and skewed distributions, enabling rigorous benchmarking of algorithms. Real-world datasets further validated the practical applicability of these methods in complex, uncontrolled environments such as healthcare, finance, cybersecurity, and natural language processing. Common evaluation metrics included accuracy, precision, recall, F1-score, area under the ROC curve (AUC-ROC), and fairness metrics, which collectively provided comprehensive performance assessments under various evolving scenarios.

From a scientific impact perspective, this survey reveals the growing influence of evolving machine learning research across both foundational and applied domains. Advances in drift detection and adaptive learning have informed the design of autonomous systems capable of long-term operation without retraining, while continual learning and fairness-aware frameworks have contributed to more responsible and explainable AI. The surveyed studies demonstrate strong cross-domain relevance, driving innovation in industrial process monitoring, medical diagnostics, predictive maintenance, and adaptive decision-making systems. These contributions reinforce the transformative potential of evolving machine learning not only for improving algorithmic robustness but also for enabling sustainable, adaptive, and ethically aligned intelligent systems.

Overall, this comprehensive survey demonstrate significant advancements and remaining challenges in evolving machine learning methods, highlighting future research opportunities particularly in integrating multiple adaptive strategies, enhancing real-time applicability, and ensuring fairness and robustness in practical, real-world applications.

\section*{Acknowledgements}

Funded by the European Union Horizon Europe programme CL2 2024-TRANSFORMATIONS-01-06 through ALFIE under Grant Agreement 101177912. Views and opinions expressed are however those of the author(s) only and do not necessarily reflect those of the European Union or the Agency. Neither the European Union nor the granting authority can be held responsible for them.

For the purpose of open access, the author(s) has (have) applied a Creative Commons Attribution (CC BY) licence to any Author Accepted Manuscript version arising from this submission. 

\bibliographystyle{plainnat}
\bibliography{refs}
\end{document}